\documentclass{article}

\PassOptionsToPackage{numbers, compress}{natbib}

\usepackage[main, final]{neurips_2026}

\usepackage[utf8]{inputenc}
\usepackage[T1]{fontenc}
\usepackage{hyperref}
\usepackage{url}
\usepackage{booktabs}
\usepackage{amsfonts}
\usepackage{nicefrac}
\usepackage{microtype}
\usepackage{xcolor}
\usepackage{graphicx}
\usepackage{subcaption}
\usepackage{wrapfig}
\usepackage{needspace}

\usepackage{amsmath}
\usepackage{amssymb}
\usepackage{mathtools}
\usepackage{amsthm}

\usepackage{algorithm}
\usepackage{algorithmic}

\usepackage[capitalize,noabbrev]{cleveref}

\usepackage{array}
\usepackage{colortbl}
\usepackage{adjustbox}
\usepackage{multirow}

\theoremstyle{plain}
\newtheorem{theorem}{Theorem}[section]

\newtheorem{lemma}[theorem]{Lemma}

\theoremstyle{definition}

\theoremstyle{remark}

\usepackage{tcolorbox}
\usepackage{listings}
\usepackage{enumitem}
\definecolor{Gray}{gray}{0.85}
\definecolor{dreamblue}{HTML}{0057d1}
\newcommand{\colorhref}[2]{{\hypersetup{pdfborder={0 0 0}}\href{#1}{\textcolor{dreamblue}{#2}}}}
\usepackage{thmtools}
\usepackage{bm}
\definecolor{table_color}{RGB}{220, 235, 255}
\newcommand{\best}[1]{\colorbox{table_color}{\textbf{#1}}}
\newcommand{\secondbest}[1]{\colorbox{black!10}{#1}}

\title{Model Immunization Beyond Condition Numbers: The Importance of Plateau Regions}

\author{%
  Quoc Nguyen$^{1}$ \quad Trung Le$^{2}$ \quad Dinh Phung$^{2}$ \quad Mehrtash Harandi$^{1}$ \\
  $^{1}$Department of Electrical and Computer Systems Engineering, Monash University, Australia\\
  $^{2}$Department of Data Science and AI, Monash University, Australia\\
  \texttt{\{minh.nguyen12, trunglm, dinh.phung, mehrtash.harandi\}@monash.edu} \\
}

\begin{document}

\maketitle

\begin{abstract}
Harmful fine-tuning can rapidly compromise safety alignment in Large Language Models, underscoring the vulnerability of deployed models to post-release modification. While prior work motivates model immunization by inducing ill-conditioned curvature to slow harmful optimization, this approach is not scalable due to the high computational cost of Hessian estimation. We refine this paradigm by extending the theoretical framework to prioritize plateau regions alongside curvature. We further introduce an immunization algorithm based on efficient Hessian estimation and bounds on the extremal singular values of the Hessian. We demonstrate the proposed immunization framework in three applications: (i) Fine-tuning-as-a-service settings where benign user data may be mixed with poisoned harmful data, (ii) open-weight settings where an adversary can fine-tune on harmful-only data with varied training dynamics, and (iii) robust unlearning, where immunized models better resist relearning attacks after unlearning. Extensive experiments confirm that our approach achieves robust defense against adversarial fine-tuning while largely maintaining general utility and downstream trainability. Our code is available \colorhref{https://github.com/minhquoc0712/BECON}{here}.
\end{abstract}

\section{Introduction}
\label{sect:intro}
Large language models (LLMs) are increasingly deployed and distributed in settings where \emph{downstream fine-tuning} is a core feature. Service providers offer fine-tuning-as-a-service (FTaaS) (\emph{e.g}., OpenAI\footnote{\url{https://platform.openai.com/docs/guides/model-optimization}}) and open-weight releases enable broad post-release adaptation. This flexibility creates a critical safety vulnerability. Even when a model is safety-aligned at release time, a small amount of harmful data can rapidly shift behavior via fine-tuning, enabling misuse and circumventing safeguards \citep{yang2023shadow,qi2024finetuning,qi2025on}.
This motivates \emph{pre-release} defenses that make harmful adaptation substantially harder.

Model immunization \citep{rosati2024immunization,rosati2024representation,tamirisa2025tamperresistant,rosati2025locking} is a framework to enable this defense by making a released model harder to fine-tune toward a target task while preserving its general capabilities and adaptation to benign tasks. In our main setting, the target task is harmful behavior, so the goal is to increase resistance to harmful fine-tuning.
Crucially, model immunization must operate under realistic threat models where an attacker can control the fine-tuning hyperparameters and use a certain training budget.

A prominent line of work formalizes immunization through \emph{loss curvature}, arguing that making the harmful objective ill-conditioned, meaning that it has a large Hessian condition number, slows gradient-based fine-tuning~\citep{rosati2025locking,zheng2025model}. While novel, the condition-number perspective can be overstated in ways that matter for practice. First, the convergence claim typically leverages a bound derived under strong convexity and specific step-size choices. As we emphasize in \cref{sec:model_immunization_beyond_condition_numbers}, a large condition number is a \emph{necessary but not sufficient} condition for slow convergence, since a slowly decreasing upper bound does not imply that the iterates cannot still make rapid progress. Second, our analysis shows that slow, harmful fine-tuning depends jointly on curvature and local steepness, since enlarging the Hessian condition number alone still allows fast loss decreases when harmful gradients remain large near the released parameters. Finally, given that estimating the largest and smallest Hessian singular values, and thus the condition number, at scale is nontrivial, existing immunization frameworks \citep{zheng2025model} neither propose an efficient method for this nor evaluate it in large models such as LLMs.

Building on the theory developed in \cref{sec:model_immunization_beyond_condition_numbers}, we argue that reliably slowing harmful optimization requires controlling \emph{both} (i) ill-conditioned curvature (large Hessian condition number $\kappa$) and (ii) \emph{plateau behavior} around the released parameters, characterized by small harmful gradient norms. Our bounds explicitly couple curvature with local gradient magnitude, highlighting that maximizing condition number alone leaves an attack surface if the harmful objective remains steep near the aligned solution. This perspective leads to a refined framework and algorithm that promotes harmful plateaus while also efficiently enlarging condition numbers.

We summarize our contributions as follows:
\begin{itemize}[leftmargin=15pt]
  \item \textbf{A refined immunization framework.} We refine the previous model immunization by condition number framework \citep{zheng2025model} by showing that convergence guarantees depend jointly on the Hessian condition number and harmful gradient norms (via tighter upper/lower bounds in \cref{sec:model_immunization_beyond_condition_numbers}). This yields a clearer characterization of when harmful fine-tuning is actually slow, and motivates \emph{plateau promotion} as a complementary immunization objective.
  \item \textbf{An efficient algorithm.} We propose \textbf{BECON}, which provides an optimization objective that (i) promotes harmful plateaus using harmful gradient perturbation and (ii) enlarges condition numbers using an efficient Hessian proxy with tractable estimates of extremal singular values, making the approach feasible for large models.
  \item \textbf{Evaluation in different applications.} We empirically demonstrate BECON's effectiveness in various applications, including increasing resistance to harmful fine-tuning in FTaaS and open-weight settings while largely retaining benign utility and trainability, and robustness against relearning attacks in unlearning settings.
\end{itemize}

\section{Related Work}
\label{sect:rwork}
\textbf{Harmful Fine-tuning Defense.} Recent research demonstrates that safety alignment can be subverted by fine-tuning models on a limited number of harmful samples \citep{yang2023shadow,yi-etal-2024-vulnerability,qi2024finetuning,huang2025booster}. In response, various defense strategies have been developed to address different stages of the model lifecycle \citep{huang2024harmful}. This paper focuses on the immunization framework, categorized as an alignment-stage defense \citep{rosati2024immunization,huang2024harmful}. Alignment-stage defenses make the model inherently resilient to malicious fine-tuning before release, and since this intervention occurs prior to fine-tuning, it applies to both open-weight models and FTaaS settings. Existing methods proactively reinforce alignment to resist harmful fine-tuning using diverse strategies. For instance, Vaccine \citep{huang2024vaccine} and Targeted-Vaccine \citep{liu2025targeted} constrain embedding shifts via perturbation-aware training, while Booster \citep{huang2025booster} and TAR \citep{tamirisa2025tamperresistant} seek safety-resilient local optima. Other approaches, such as RMU \citep{li2024the}, CTRAP \citep{yi2025ctrap}, and RepNoise \citep{rosati2024representation}, utilize unlearning and noise approximation to erase or hinder training on harmful knowledge. Furthermore, VAA \citep{chen2025vulnerabilityaware} enhances safety alignment by identifying and upweighting vulnerable alignment data, as a subset of alignment data is more prone to forgetting, and LoX \citep{perin2025lox} utilizes a training-free method to reinforce safety by extrapolating the safety subspace of an aligned LLM.

\textbf{Relearning Attacks in LLM Unlearning.} Recent studies \citep{lynch2024eight, hu2025unlearning, zhang2025catastrophic, shumailov2024ununlearning, barez2025open, deeb2024unlearning} demonstrate that LLMs subjected to unlearning algorithms can recover forgotten knowledge after fine-tuning on the original data or implicitly related samples.
Mechanistic studies \citep{deeb2024unlearning, che2024model} reveal that many unlearning methods remain vulnerable to recovery via fine-tuning, suggesting that they often suppress or hide access to the targeted knowledge rather than removing it from the model weights.
Recent work \citep{fan2025towards} has proposed a solution based on Sharpness-Aware Minimization (SAM) \citep{foret2021sharpnessaware}, which improves robustness by enforcing the unlearned model to converge to a flat region of the loss landscape, thereby hindering the optimization process of relearning attacks.

\textbf{Model Immunization with Condition Number.} Prior work has primarily studied the condition number as a measure of numerical stability and optimization robustness. In this line of research, minimizing the condition number has been shown to improve training stability and robustness to noise~\citep{balazs2024trainable, nenov2024almost}. In contrast, recent studies on \emph{model immunization} deliberately take the opposite approach: they seek to \emph{maximize} the Hessian condition number of unsafe objectives in order to hinder or slow harmful fine-tuning. One line of work introduces a framework based on this condition number maximization strategy and validates it using linear and vision architectures \citep{zheng2025model}. Concurrent work explores a similar objective by directly increasing the largest singular values of the model matrices to increase the lower bound of the Hessian's maximum singular value \citep{rosati2025locking}. Building on these foundations, our proposed method improves the theoretical aspects of condition-number-based immunization \citep{zheng2025model} and extends its practicality with an efficient algorithm applicable to large-scale settings such as LLMs. Complementarily, \citet{rosati2026limits} establishes that purely spectral-based defenses face fundamental limitations, a finding consistent with our analysis showing that resistance depends jointly on the condition number and harmful loss plateaus.

\section{Approach}
\label{sect:background}
\subsection{Condition Numbers and Model Immunization}

Immunization via the condition number of the Hessian is rooted in classical convergence results for gradient-based optimization \cite{zheng2025model}. The key idea in~\cite{zheng2025model} builds on a well-known result from optimization theory. Consider the optimization problem $\min_{\bm{\theta}}\mathcal{L}\left(\bm{\theta}\right)$ where $\mathcal{L}$ is a strongly convex function and has a Hessian $\bm{H}(\bm{\theta}) = \nabla^2 \mathcal{L}(\bm{\theta})$. Let $\sigma_{\min}$ and $\sigma_{\max}$ denote the minimum and maximum singular values of the Hessian evaluated at the current model parameters, with the parameter dependence suppressed for readability. The constant step-size steepest descent algorithm has the following convergence rate \citep{bubeck2015convex}:
\begin{equation}
\Vert\bm{\theta}_{t}-\bm{\theta}^{*}\Vert^{2}\leq\left(1-\frac{\sigma_{\min}}{\sigma_{\max}}\right)^{t}\Vert\bm{\theta}_{0}-\bm{\theta}^{*}\Vert^{2},\label{eq:cond_convergence_rate}
\end{equation}
where $\bm{\theta}^*$ denotes the optimal solution, and $\bm{\theta}_t$ denotes the steepest descent iterate at step $t$.

Defining the condition number $\kappa(\bm{H}) \coloneqq \frac{\sigma_{\max}}{\sigma_{\min}}$, the strong convexity assumption ensures a strictly positive lower curvature bound, so $\sigma_{\min}>0$ and $\kappa(\bm{H})$ is well-defined. A large $\kappa(\bm{H})$ implies the convergence factor $1 - 1 / \kappa \approx 1$, potentially leading to very slow convergence as discussed in \cite{zheng2025model}. While this observation is compelling, it should be interpreted with caution: \textit{Since \eqref{eq:cond_convergence_rate} is an upper bound, a slow decrease of the bound does not strictly imply slow convergence of the iterates.} Thus, a high condition number is a necessary but not sufficient indicator for poor optimization performance, motivating the need for more refined theoretical analysis.

 \subsection{Model Immunization Beyond Condition Numbers}
 \label{sec:model_immunization_beyond_condition_numbers}
 We consider the objective function $\mathcal{L}(\bm{\theta})$ (i.e., the empirical loss over a training set), which satisfies the following assumptions:

\textbf{(A1)} The objective function $\mathcal{L}(\bm{\theta})$ has the Hessian matrix $\bm{H}$ satisfying: $\sigma_{\min}\bm{I} \prec \bm{H} \prec \sigma_{\max}\bm{I}$ for the positive min/max singular values $\sigma_{\min}/\sigma_{\max}$.

\textbf{(A2)} There exists a global minima: $\bm{\theta}^{*}=\text{argmin}_{\bm{\theta}}\mathcal{L}\left(\bm{\theta}\right)$ and $\mathcal{L}^{*}=\mathcal{L}\left(\bm{\theta}^{*}\right)$.

We first prove that the Polyak–Lojasiewicz (PL) inequality holds for $\sigma_{\min}$ in the following lemma.
\begin{lemma} \label{PL_inequality}
We have the following inequalities

(i) The PL-typed inequality holds for $\sigma_{\min}$:
\begin{equation}
\frac{1}{2}\Vert\nabla\mathcal{L}\left(\bm{\theta}\right)\Vert^{2}\geq\sigma_{\min}\left(\mathcal{L}\left(\bm{\theta}\right)-\mathcal{L}^{*}\right), \quad \forall\, \bm{\theta},\label{eq:kl_min}
\end{equation}
where $\mathcal{L}^{*}=\min_{\bm{\theta}}\mathcal{L}\left(\bm{\theta}\right)$.  

(ii) The opposite PL-type inequality holds for $\sigma_{\max}$:
\begin{equation}
\frac{1}{2}\Vert\nabla\mathcal{L}\left(\bm{\theta}\right)\Vert^{2}\leq\sigma_{\max}\left(\mathcal{L}\left(\bm{\theta}\right)-\mathcal{L}^{*}\right), \quad \forall\, \bm{\theta.}\label{eq:kl_max}
\end{equation}

\end{lemma}

Denote $\bm{\theta}_t$ as the solution (i.e., the model) at the time step $t$. Assuming that we use gradient descent to update a new model: $\bm{\theta}_{t+1}=\bm{\theta}_{t}-\eta\nabla\mathcal{L}\left(\bm{\theta}_{t}\right)$ with the learning rate $\eta>0$, the following lemma specifies the range of the learning rate $\eta$ for efficient and useful updates.
\begin{lemma}\label{lem:eta_range}
The update $\bm{\theta}_{t+1}=\bm{\theta}_{t}-\eta\nabla\mathcal{L}\left(\bm{\theta}_{t}\right)$ is efficient in reducing the new objective $\mathcal{L}(\bm{\theta}_{t+1})$ (i.e., $\mathcal{L}\left(\bm{\theta}_{t+1}\right)<\mathcal{L}\left(\bm{\theta}_{t}\right)$) if $0< \eta < \frac{2}{\sigma_{\max}}$.    
\end{lemma}

In what follows, we develop necessary inequalities to indicate the quantities that slow down the convergence of the OP of interest, which can be applied directly to the model immunization. 
\begin{theorem}\label{thm:main}
Considering the learning rate $\eta \in \left(0, \frac{2}{\sigma_{\max}}\right)$, we have the following inequalities:

(i) We can upper-bound $\mathcal{L}(\bm{\theta}_{t+1}) - \mathcal{L}^*$ by
\begin{equation}
\mathcal{L}\left(\bm{\theta}_{t+1}\right)-\mathcal{L}^{*}\leq\left(1-\frac{1}{\kappa}\right)^{t+1}\left(\mathcal{L}\left(\bm{\theta}_{0}\right)-\mathcal{L}^{*}\right)+\frac{\left(\sigma_{\max}\eta-1\right)^{2}}{2\sigma_{\max}}\sum_{i=1}^{t}\left(1-\frac{1}{\kappa}\right)^{t-i}\Vert\nabla\mathcal{L}\left(\bm{\theta}_{i}\right)\Vert^{2}.\label{eq:upper_bound}
\end{equation}
(ii) We can lower-bound $\mathcal{L}(\bm{\theta}_{t+1}) - \mathcal{L}^*$ by 

$\square$ If $\kappa > 2$ and $\eta\in\left[\frac{1}{\kappa\sigma_{\max}},\frac{2}{\sigma_{\max}}\right)$, we have the lower bound
\begin{equation}
\begin{aligned}
\mathcal{L}\left(\bm{\theta}_{t+1}\right)-\mathcal{L}^{*}
&\geq \left(1-\frac{1}{\kappa}\right)^{3\left(t+1\right)}
\left(\mathcal{L}\left(\bm{\theta}_{0}\right)-\mathcal{L}^{*}\right) \\
&\quad - \left(\eta-\frac{1}{\kappa\sigma_{\max}}\right)
\left(1-\frac{1}{\kappa^{2}}\right)
\sum_{i=1}^{t}\left(1-\frac{1}{\kappa}\right)^{3\left(t-i\right)}
\Vert\nabla\mathcal{L}\left(\bm{\theta}_{i}\right)\Vert^{2}.
\end{aligned}
\label{eq:lower_bound1}
\end{equation}
$\square$ If the learning rate $0<\eta<\frac{1}{\kappa\sigma_{\max}}$, we have the lower bound
\begin{equation}
\mathcal{L}\left(\bm{\theta}_{t+1}\right)-\mathcal{L}^{*}\geq\left(1-\frac{1}{\kappa}\right)^{t+1}
\hspace{-3mm}
\left(\mathcal{L}\left(\bm{\theta}_{0}\right)-\mathcal{L}^{*}\right)-\frac{\eta}{2}\left(1-\frac{\sigma_{\min}\eta}{2}\right)\sum_{i=1}^{t}\left(1-\frac{1}{\kappa}\right)^{t-i}\Vert\nabla\mathcal{L}\left(\bm{\theta}_{i}\right)\Vert^{2}.\label{eq:lower_bound2}
\end{equation}
\end{theorem}

Theorem \ref{thm:main} develops the lower and upper bounds for the \textit{convergence gap} $\mathcal{L}\left(\bm{\theta}_{t+1}\right)-\mathcal{L}^{*}$. From the lower bounds in (\ref{eq:lower_bound1}) and (\ref{eq:lower_bound2}), we observe that to increase the \textit{convergence gap} for slowing down convergence, we need to (i) minimize $\kappa^{-1}=\frac{\sigma_{\min}}{\sigma_{\max}}$ (or maximize the condition number $\kappa=\frac{\sigma_{\max}}{\sigma_{\min}}$) and (ii) minimize the gradient norms $\Vert\nabla\mathcal{L}\left(\bm{\theta}_{i}\right)\Vert, i=1,...,t$. Moreover, according to the upper bound (\ref{eq:upper_bound}), when minimizing the gradient norms, the upper bound is closer to the lower bound, implying a tighter approximation. Additionally, by setting the learning rate $\eta = \frac{1}{\sigma_{\max}}$, our upper bound reduces to the bound in (\ref{eq:cond_convergence_rate}), indicating that our upper bound is more general than the one in (\ref{eq:cond_convergence_rate}), which only considers the constant learning rate $\eta = \frac{1}{\sigma_{\max}}$. The proofs of these lemmas and theorems are presented in Appendix \ref{app:all_proof}. Finally, we provide a discussion in Appendix~\ref{app:condition_number_discussion} of why small harmful gradient norms do not imply that the model is close to the harmful optimum.

\section{Proposed Method}
\label{sect:method}
Our primary objective is to immunize LLMs against adaptation to harmful datasets in particular, and to undesired target datasets in general. Specifically, we aim to configure the safety alignment phase such that downstream users cannot effectively fine-tune the model on harmful datasets $\mathcal{D}_{\mathrm{H}}$ after the aligned model is released for fine-tuning. Motivated by the theoretical insights in \cref{sec:model_immunization_beyond_condition_numbers}, our objective has two complementary parts: harmful plateau promotion and condition number enlargement.

\textbf{Harmful plateau promotion.} Following \cref{thm:main}, we encourage \emph{plateau behavior} on harmful data, so that a single step of harmful fine-tuning yields only a small decrease in harmful loss. Let $\mathcal{L}_{\mathrm{H}}(\bm{\theta})$ be the harmful loss, and define the $\ell_2$ ball $\mathcal{B}_{\rho}(\bm{\theta})\!:=\!\{\bm{\phi}\!:\!\|\bm{\phi}-\bm{\theta}\|_2\le \rho\}$. Intuitively, the plateau objective penalizes how much the harmful loss can decrease under a small update within this ball. We approximate this local harmful improvement by a single normalized harmful-gradient step, which lies on the boundary of $\mathcal{B}_{\rho}(\bm{\theta})$:
\begin{equation}
\bm{\theta}_{\text{pert}}\coloneqq\bm{\theta}-\rho\,\nabla_{\bm{\theta}} \mathcal{L}_{\mathrm{H}}(\bm{\theta})\big/\|\nabla_{\bm{\theta}} \mathcal{L}_{\mathrm{H}}(\bm{\theta})\|_2.
\end{equation}
If the harmful landscape is locally flat along this attack direction, then moving to $\bm{\theta}_{\text{pert}}$ yields little improvement in harmful loss. For small $\rho$, we can explicitly derive the connection between this one-step improvement and gradient suppression via a first-order Taylor expansion:
\begin{equation}
\mathcal{L}_{\mathrm{H}}(\bm{\theta}_{\text{pert}}) \approx \mathcal{L}_{\mathrm{H}}(\bm{\theta}) + \nabla_{\bm{\theta}} \mathcal{L}_{\mathrm{H}}(\bm{\theta})^\top (\bm{\theta}_{\text{pert}} - \bm{\theta}) = \mathcal{L}_{\mathrm{H}}(\bm{\theta}) - \rho \frac{\|\nabla_{\bm{\theta}} \mathcal{L}_{\mathrm{H}}(\bm{\theta})\|_2^2}{\|\nabla_{\bm{\theta}} \mathcal{L}_{\mathrm{H}}(\bm{\theta})\|_2} = \mathcal{L}_{\mathrm{H}}(\bm{\theta}) - \rho \|\nabla_{\bm{\theta}} \mathcal{L}_{\mathrm{H}}(\bm{\theta})\|_2.
\end{equation}
Thus, discouraging this one-step harmful loss decrease directly promotes small harmful gradients.

In practice, per-sample losses within a harmful minibatch can vary significantly. To ensure the plateau objective prioritizes samples that are most sensitive to the harmful update, we weight examples by the change in their losses under the perturbation. For a harmful minibatch $\{\bm{x}_i\}_{i=1}^{B}$ with per-sample losses $\ell_i(\bm{\theta})$, define:
\begin{equation}
\Delta_i \coloneqq \left|\ell_i(\bm{\theta})-\ell_i(\bm{\theta}_{\text{pert}})\right|,
\qquad
w_i \;\coloneqq\; \frac{\exp\!\big(\Delta_i/\tau\big)}{\sum_{j=1}^{B}\exp\!\big(\Delta_j/\tau\big)},
\end{equation}
where $\tau>0$ is the softmax temperature, controlling how sharply we emphasize the most sensitive harmful samples. We then instantiate the plateau loss as:
\begin{equation}
\mathcal{L}_{\text{plateau}}(\bm{\theta}) \coloneqq \sum_{i=1}^{B} w_i\left(\ell_i(\bm{\theta})-\ell_i(\bm{\theta}_{\text{pert}})\right).
\end{equation}

\textbf{Condition number enlargement.} Complementary to the plateau objective, we distort the harmful loss geometry by increasing the condition number $\kappa(\bm{H}) = \sigma_{\max}/\sigma_{\min}$ of the harmful Hessian. However, 
computing the explicit Hessian $\bm{H}$ is intractable for LLMs. Inspired by Sophia \citep{liu2024sophia}, we use the Gauss-Newton approximation.
Let $\bm{f}(\bm{\theta}, \bm{x}) \in \mathbb{R}^V$ be the logits (pre-softmax scores) and $\ell(\bm{f}(\bm{\theta}, \bm{x}), \bm{y})$ be the cross-entropy loss.
The Hessian of the loss can be approximated by the Gauss-Newton matrix:
\begin{equation}
\nabla_{\bm{\theta}}^{2} \ell(\bm{\theta})
\approx \bm{J}_{\bm{\theta}} \bm{f}(\bm{\theta}, \bm{x}) \cdot \bm{S} \cdot \bm{J}_{\bm{\theta}} \bm{f}(\bm{\theta}, \bm{x})^{\top},
\end{equation}
where $\bm{J}_{\bm{\theta}} \bm{f}(\bm{\theta}, \bm{x}) \in \mathbb{R}^{d \times V}$ is the Jacobian of the logits with respect to parameters $\bm{\theta}$ and
$\bm{S} \in \mathbb{R}^{V \times V}$ is the Hessian of the loss with respect to the logits.
Here, we omit the residual term involving the second derivative of the mapping $f$ as it is often negligible \citep{liu2024sophia}.
The Gauss-Newton term can be expressed as the covariance of gradients computed with labels $\hat{\bm{y}}$ sampled from the model's predictive distribution:
\begin{align}
\bm{H}_{\mathrm{GN}} &\coloneqq \bm{J}_{\bm{\theta}} \bm{f}(\bm{\theta}, \bm{x}) \cdot \bm{S} \cdot \bm{J}_{\bm{\theta}} \bm{f}(\bm{\theta}, \bm{x})^{\top} \nonumber \\
&= \mathbb{E}_{\hat{\bm{y}}\sim p_{\bm{\theta}}(\cdot|\bm{x})}[\nabla \ell(\bm{f}(\bm{\theta}, \bm{x}), \hat{\bm{y}}) \nabla \ell(\bm{f}(\bm{\theta}, \bm{x}), \hat{\bm{y}})^\top].
\end{align}
This identity allows us to represent curvature using a Gram matrix built from sub-minibatch gradients sampled from the model.

\begin{theorem}
\label{thm:hessian_estimation_ce}
Consider a harmful minibatch of size $B$ partitioned into $K$ sub-minibatches of size $M$ ($B=KM$). Let $\bm{g}_k$ be the average gradient of the $k$-th sub-minibatch computed with labels sampled from the model's predictive distribution. Define $\bm{G}=[\bm{g}_1,\dots,\bm{g}_K]\in\mathbb{R}^{d\times K}$. The Gram matrix estimator of the minibatch Gauss-Newton curvature is given by:
\begin{equation}
\widehat{\bm{H}}_{\mathrm{GN}} \coloneqq \frac{M^2}{B} \bm{G} \bm{G}^\top.
\end{equation}
\end{theorem}
The proof of this theorem is provided in Appendix~\ref{app:all_proof}. In our implementation, the sub-minibatch gradients are computed with the ground-truth harmful labels instead of sampled labels (see Appendix~\ref{app:condition_number_discussion} for a discussion and a comparison with sampled labels).

\noindent With this estimator, each sub-minibatch gradient norm $\|\bm{g}_k\|_2^2$ becomes a cheap directional curvature proxy.
To connect $\|\bm{g}_k\|_2^2$ to extremal curvature, note that $\widehat{\bm{H}}_{\mathrm{GN}} = \tfrac{M^2}{B}\bm{G}\bm{G}^\top$ is positive semidefinite and $\bm{G}\bm{G}^\top$ and $\bm{G}^\top \bm{G}$ share the same non-zero eigenvalues. For the $k$-th standard basis vector $\bm{e}_k\in\mathbb{R}^K$,
\begin{equation}
\bm{e}_k^\top (\bm{G}^\top \bm{G}) \bm{e}_k \;=\; \|\bm{g}_k\|_2^2.
\end{equation}
Hence, by Rayleigh quotient bounds \citep{horn2012matrix},
\begin{equation}
\sigma_{\max}(\widehat{\bm{H}}_{\mathrm{GN}}) \;\ge\; \tfrac{M^2}{B}\max_k \|\bm{g}_k\|_2^2, \qquad \sigma_{\min}(\widehat{\bm{H}}_{\mathrm{GN}}) \;\le\; \tfrac{M^2}{B}\min_k \|\bm{g}_k\|_2^2,
\end{equation}
where $\sigma_{\min}$ denotes the smallest non-zero singular value. Dropping the constant factor $\tfrac{M^2}{B}$ yields simple computable proxies, and we maximize the condition number by simultaneously decreasing the smallest singular value proxy and increasing the largest:
\begin{equation}
\mathcal{L}_{\text{cn-min}}(\bm{\theta}) \coloneqq \min_{k} \|\bm{g}_k\|_2^2, \qquad \mathcal{L}_{\text{cn-max}}(\bm{\theta}) \coloneqq \max_{k} \|\bm{g}_k\|_2^2.
\label{eq:cn_proxies}
\end{equation}
Minimizing $\mathcal{L}_{\text{cn-min}}$ suppresses the curvature in the direction of least variation by lowering the upper bound of $\sigma_{\min}$, while maximizing $\mathcal{L}_{\text{cn-max}}$ amplifies the curvature in the direction of greatest variation by raising the lower bound of $\sigma_{\max}$.

Optimizing these curvature proxies requires differentiating through the gradient norm $\|\bm{g}\|_2^2$, which involves the Hessian-vector product $\bm{H}\bm{g}$. To avoid instantiating the Hessian, we employ a finite-difference approximation. For a selected sub-minibatch gradient $\bm{g}$, we perturb the weights by $\varepsilon$ to obtain $\bm{\theta}' = \bm{\theta} - \varepsilon \bm{g}$. The gradient of the squared norm is then approximated as:
\begin{equation}
\nabla_{\bm{\theta}} \|\bm{g}\|_2^2 = 2\bm{H}\bm{g} \approx \frac{2}{\varepsilon} \big( \nabla_{\bm{\theta}}\mathcal{L}(\bm{\theta}) - \nabla_{\bm{\theta}}\mathcal{L}(\bm{\theta}') \big).
\end{equation}
In practice, for each harmful minibatch, we compute the set $\{\bm{g}_k\}$ and identify the indices $k_{\min}$ and $k_{\max}$ corresponding to the smallest and largest norms. We then perform the finite-difference calculation only for these two selected directions. This ensures the computational overhead is limited to two additional backward passes per step to make the method scalable to large models.

\textbf{Overall formulation.} Let $\mathcal{D}_{\mathrm{align}}$ denote the alignment dataset consisting of harmful requests paired with safe responses. We define $\mathcal{L}_{\text{align}}$ as the standard cross-entropy loss on $\mathcal{D}_{\mathrm{align}}$. Putting everything together, the overall optimization objective of BECON is
\begin{equation}
\mathcal{L}_{\text{align}}(\bm{\theta})
\;+\; \lambda_1\,\mathcal{L}_{\text{plateau}}(\bm{\theta})
\;+\; \lambda_2\,\mathcal{L}_{\text{cn-min}}(\bm{\theta})
\;-\; \lambda_3\,\mathcal{L}_{\text{cn-max}}(\bm{\theta}),
\label{eq:overall_formulation}
\end{equation}
where the hyperparameters $\lambda_1$, $\lambda_2$, and $\lambda_3$ balance the contributions of the different terms. In our implementation, when computing $\mathcal{L}_{\text{plateau}}$ and the condition-number losses, we treat the perturbed parameters (e.g., $\bm{\theta}_{\text{pert}}$ and $\bm{\theta}'=\bm{\theta}-\varepsilon \bm{g}$) as stop-gradient quantities and do not backpropagate through the perturbation construction. We summarize the full BECON training procedure in \cref{alg:algorithm}, provide further comparison with Booster~\citep{huang2025booster} and Antibody~\citep{nguyen2026antibody} in \cref{app:further_discussion}, and discuss the strength of our condition-number surrogate relative to prior work \cite{zheng2025model} in \cref{app:condition_number_discussion}.

\section{Experiments}
In this section, we demonstrate the effectiveness of our immunization approach in three settings. First, we study the FTaaS setup, where the service provider controls fine-tuning. Second, we consider the open-weight setting, where adversaries can fine-tune the model with different training dynamics. Finally, we show that our method provides robust unlearning against relearning attacks.

\subsection{Defense against harmful fine-tuning in FTaaS}
\textbf{Experiment Setup \& Threat Model.} We evaluate our framework in the FTaaS setting, simulating a two-stage pipeline: \textit{alignment} and \textit{fine-tuning}. In the alignment stage, the provider ensures safety using a safety dataset $\mathcal{D}_{\text{align}}$ (refusals) and a harmful dataset $\mathcal{D}_{\text{harm}}$ (harmful Q\&A). Subsequently, in the fine-tuning stage, a user submits a dataset $\mathcal{D}_{\text{task}}$ containing $n$ samples. Following prior work \citep{qi2024finetuning,huang2025booster}, we assume an attack scenario where the user injects harmful samples into this dataset, such that $\mathcal{D}_{\text{task}}$ consists of $(1-p)\%$ benign task data and $p\%$ harmful data. The defender's goal is to maintain safety without degrading performance on the benign task.

\textbf{Datasets and Models.} We utilize harmful data curated from the BeaverTails dataset \citep{ji2023beavertails}. For benign tasks, we employ SST2 \citep{socher2013recursive}, AGNEWS \citep{zhang2015character} and GSM8K \citep{cobbe2021training}. We evaluate performance across three model families: Llama-2-7B \citep{touvron2023llama}, Qwen-2-7B \citep{yang2024qwen2} and Gemma-2-9B \citep{team2024gemma}. We compare our approach against standard Supervised Fine-Tuning (SFT) and other defense baselines: Vaccine \citep{huang2024vaccine}, Booster \citep{huang2025booster}, and Lisa \citep{huang2024lisa}.

\textbf{Evaluation Metrics.} Following the evaluation protocol of prior works \citep{huang2024harmful}, we assess models on two metrics: (i) Safety, measured by the Harmful Score (HS), which is the percentage of responses to malicious prompts flagged as harmful by a moderation model \citep{ji2023beavertails}, and (ii) Fine-tuning performance, measured by Fine-tuning Accuracy (FA) on the benign tasks.

\textbf{Implementation Details.} We utilize LoRA \citep{hu2022lora} for efficient training. Unless otherwise stated, our default setting uses Llama-2-7B fine-tuned on the GSM8K dataset ($n=1000$) with a harmful injection ratio of $p=0.2$. Results are averaged over three random seeds. Full hyperparameters and training configurations are detailed in Appendix \ref{app:exp_details_ftaas}.

\begin{table*}[!h]
\centering
\caption{Performance of models trained on different fine-tuning datasets. The best and the second best are highlighted in \colorbox[RGB]{220, 235, 255}{blue} and \secondbest{gray}, respectively. }
\label{tab:experiment_different_model_dataset}
\begin{adjustbox}{width=0.99\textwidth}
\begin{tabular}{l|cccccc|cccccc|cccccc}
\toprule
 & \multicolumn{6}{c|}{\textbf{Llama-2-7B}} & \multicolumn{6}{c|}{\textbf{Gemma2-9B}} & \multicolumn{6}{c}{\textbf{Qwen2-7B}} \\
\cmidrule(lr){2-7} \cmidrule(lr){8-13} \cmidrule(lr){14-19}
Methods & \multicolumn{2}{c}{\textbf{SST2}} & \multicolumn{2}{c}{\textbf{AGNEWS}} & \multicolumn{2}{c}{\textbf{GSM8K}} & \multicolumn{2}{c}{\textbf{SST2}} & \multicolumn{2}{c}{\textbf{AGNEWS}} & \multicolumn{2}{c}{\textbf{GSM8K}} & \multicolumn{2}{c}{\textbf{SST2}} & \multicolumn{2}{c}{\textbf{AGNEWS}} & \multicolumn{2}{c}{\textbf{GSM8K}} \\
\cmidrule(lr){2-3} \cmidrule(lr){4-5} \cmidrule(lr){6-7} \cmidrule(lr){8-9} \cmidrule(lr){10-11} \cmidrule(lr){12-13} \cmidrule(lr){14-15} \cmidrule(lr){16-17} \cmidrule(lr){18-19}
 & HS $\downarrow$ & FA $\uparrow$ & HS $\downarrow$ & FA $\uparrow$ & HS $\downarrow$ & FA $\uparrow$ & HS $\downarrow$ & FA $\uparrow$ & HS $\downarrow$ & FA $\uparrow$ & HS $\downarrow$ & FA $\uparrow$ & HS $\downarrow$ & FA $\uparrow$ & HS $\downarrow$ & FA $\uparrow$ & HS $\downarrow$ & FA $\uparrow$ \\
\midrule
SFT        & 36.29 & \secondbest{92.70} & 34.57 & 85.40 & 23.94 & 10.90 & 40.77 & 94.11 & 39.91 & 84.47 & 32.05 & 56.73 & 33.86 & 93.31 & 31.00 & \secondbest{85.77} & 14.54 & 64.66 \\
Vaccine    & 44.16 & 91.71 & 39.73 & 82.43 & 23.60 & 11.70 & 51.65 & \best{94.50} & 51.64 & \best{85.27} & 38.24 & 54.37 & 29.80 & 93.22 & 28.66 & 84.93 & 18.17 & 62.20 \\
Lisa       & 22.94 & 92.51 & 20.93 & 84.50 & \secondbest{5.86}  & 9.23  & 34.67 & 94.04 & 33.33 & 83.80 & \secondbest{13.02} & 54.97 & 18.46 & 92.85 & 15.26 & 85.07 & 3.91  & 63.90 \\
Booster    & \secondbest{14.31} & 92.59 & \secondbest{15.88} & \secondbest{86.70} & 9.06  & \best{16.27} & \secondbest{28.47} & 94.15 & \secondbest{19.22} & 84.30 & 22.13 & \best{58.97} & \secondbest{3.96}  & \secondbest{94.19} & \secondbest{3.48}  & 85.13 & \best{2.19}  & \best{68.63} \\
BECON       & \best{11.16} & \best{93.15} & \best{13.59} & \best{86.77} & \best{3.72} & \secondbest{15.43} & \best{16.21} & \secondbest{94.42} & \best{4.24}  & \secondbest{84.80}  & \best{8.15}  & \secondbest{58.07} & \best{3.34}  & \best{94.76} & \best{2.91}  & \best{86.00} & \secondbest{2.48}  & \secondbest{67.93} \\
\bottomrule
\end{tabular}
\end{adjustbox}
\end{table*}

\textbf{Generalization to different fine-tuning datasets.} As shown in \cref{tab:experiment_different_model_dataset}, BECON provides the strongest overall defense across datasets and model families. It achieves the lowest HS in eight of nine settings, and ranks second on Qwen2-7B with GSM8K, where its HS is 2.48\% compared to 2.19\% for Booster. BECON also preserves downstream trainability, attaining the best or second-best FA in every setting. These results show that BECON reliably mitigates harmful fine-tuning while remaining competitive on task performance. Additional FTaaS experiments on the \textbf{more recent Qwen-3-4B model} \citep{yang2025qwen3}, under \textbf{harmful data distribution shift}, and for the \textbf{local safety loss landscape} are reported in Appendix \ref{app:exp_qwen3}, \ref{app:harmful_data_shift}, and \ref{app:safety_loss_landscape}, respectively.

\begin{table*}[!h]
\centering
\caption{Performance under different harmful ratios in the default setting. \textit{clean} means no harmful samples or $p=0$. The best and the second best are highlighted in \colorbox[RGB]{220, 235, 255}{blue} and \secondbest{gray}, respectively.}
\label{tab:experiment_varying_harmful_ratio}
\begin{adjustbox}{max width=0.99\textwidth}
\begin{tabular}{l|ccccccc|ccccccc}
\toprule
    Methods &   \multicolumn{7}{c}{ \textbf{Harmful Score}  $\downarrow$}& \multicolumn{7}{c}{\textbf{Fine-tuning Accuracy}  $\uparrow$}\\
        \cmidrule(lr){2-8} \cmidrule(lr){9-15}
$(n=1000)$ & clean & $p=0.05$ & $p=0.1$ & $p=0.15$ & $p=0.2$ & $p=0.25$ & Average & clean & $p=0.05$ & $p=0.1$ & $p=0.15$ & $p=0.2$ & $p=0.25$ & Average \\
\midrule
SFT          & 2.05 & 10.68 & 15.59 & 19.06 & 23.94 & 27.85 & 16.53 & 12.23 & 11.83 & 11.43 & 10.90 & 10.90 & 10.53 & 11.30 \\
Vaccine      & \secondbest{1.29} & 7.15 & 13.07 & 19.55 & 23.60 & 29.33 & 15.67 & 12.27 & 12.23 & 11.77 & 11.77 & 11.77 & 11.77 & 11.93 \\
Lisa         & \best{0.95} & 3.05 & 3.82 & \secondbest{4.62} & \secondbest{5.86} & \secondbest{8.16} & \secondbest{4.41} & 9.97 & 9.57 & 9.60 & 8.90 & 9.23 & 9.20 & 9.41  \\
Booster      & 1.38 & \best{1.76} & \secondbest{2.91} & 5.10 & 9.06 & 13.49 & 5.62 & \best{16.53} & \secondbest{15.97} & \best{16.50} & \best{16.30} & \best{16.43} & \best{15.70} & \best{16.24} \\
BECON         & 1.86 & \secondbest{2.00} & \best{2.20} & \best{3.62} & \best{3.72} & \best{7.06} & \best{3.41} & \secondbest{15.97} & \best{16.10} & \secondbest{15.13} & \secondbest{15.20} & \secondbest{15.43} & \secondbest{15.57} & \secondbest{15.57} \\
\bottomrule
\end{tabular}
\end{adjustbox}
\end{table*}

\textbf{Robustness across different harmful ratios.} As shown in \cref{tab:experiment_varying_harmful_ratio}, BECON remains robust as the harmful ratio \(p\) increases. It achieves the best average HS of 3.41\% and the lowest HS for all \(p \geq 0.1\), while avoiding the utility loss of Lisa. Although Booster attains the highest average FA, its safety degrades more quickly under heavier poisoning, reaching 13.49\% HS at \(p=0.25\) compared to 7.06\% for BECON. Overall, BECON provides the strongest safety-utility trade-off across poisoning ratios.

\label{sec:ftaas}

\subsection{Immunization against harmful fine-tuning attacks in open-weight settings}
\label{sec:open_weight_immunization}
In this section, we study \emph{open-weight immunization}, a scenario where an immunized checkpoint is released, and an adversary possesses full control over post-release fine-tuning. Concretely, we evaluate robustness against \emph{harmful-only} fine-tuning using curated harmful instructions from BeaverTails \citep{ji2023beavertails}. To rigorously simulate this threat, we stress-test the model by sweeping across the attacker's learning rate and the number of fine-tuning epochs; these results are reported in \cref{tab:experiment_varying_epochs,tab:experiment_varying_learning_rate}. We perform these evaluations on Llama-2-7B \citep{touvron2023llama}, where all fine-tuning is conducted using LoRA \citep{hu2022lora} with the AdamW optimizer. Further implementation details are provided in Appendix \ref{app:exp_details_immunization}, and computational overhead analysis is reported in Appendix \ref{app:system_evaluation}. We further evaluate \textbf{full-parameter fine-tuning attacks} in Appendix \ref{app:full_finetuning}.

Following the protocols established by \cite{rosati2024immunization} and \cite{zheng2025model}, we assess performance across three dimensions, comparing our method against existing open-weight defenses such as TAR \citep{tamirisa2025tamperresistant} and RepNoise \citep{rosati2024representation}. Specifically, we evaluate: (i) harmful fine-tuning resistance, measured using the Harmful Score on the BeaverTails-Evaluation split; (ii) utility preservation, which assesses the retention of helpful behavior via accuracy on MMLU \citep{hendrycks2021measuring}, TruthfulQA \citep{lin2022truthfulqa}, HellaSwag \citep{zellers-etal-2019-hellaswag}, and ARC-Easy \citep{clark2018think}; and (iii) trainability, which evaluates the model's ability to learn from harmless datasets via downstream fine-tuning accuracy on SST2 \citep{socher2013recursive}, AGNEWS \citep{zhang2015character}, and GSM8K \citep{cobbe2021training}.

\begin{table}[!h]
\centering
\caption{Harmful immunization performance across training epochs. The best and the second best are highlighted in \colorbox[RGB]{220, 235, 255}{blue} and \secondbest{gray}, respectively.}
\label{tab:experiment_varying_epochs}
\begin{adjustbox}{max width=0.80\columnwidth}
\begin{tabular}{l|cccccc}
\toprule
    Methods & \multicolumn{6}{c}{\textbf{Harmful Score} $\downarrow$} \\
    \cmidrule(lr){2-7}
    & Epoch=5 & Epoch=10 & Epoch=20 & Epoch=50 & Epoch=75 & Epoch=100 \\
\midrule
TAR      & 42.63 & 50.64 & 54.51 & 55.08 & 55.79 & \secondbest{54.94} \\
RepNoise & \secondbest{13.02} & \secondbest{16.31} & \best{15.74} & \best{43.35} & \secondbest{53.51} & 55.79 \\
BECON    & \best{1.00} & \best{3.43} & \secondbest{27.47} & \secondbest{50.21} & \best{50.50} & \best{52.50} \\
\bottomrule
\end{tabular}
\end{adjustbox}
\end{table}

\begin{table}[!h]
\centering
\caption{Harmful immunization performance across different learning rates. The best and the second best are highlighted in \colorbox[RGB]{220, 235, 255}{blue} and \secondbest{gray}, respectively.}
\label{tab:experiment_varying_learning_rate}
\begin{adjustbox}{max width=0.67\columnwidth}
\begin{tabular}{l|cccccc}
\toprule
    Methods & \multicolumn{6}{c}{\textbf{Harmful Score} $\downarrow$} \\
    \cmidrule(lr){2-7}
    & lr=5e-6 & lr=1e-5 & lr=5e-5 & lr=1e-4 & lr=5e-4 & lr=1e-3 \\
\midrule
TAR                          & 28.47 & 42.63 & 56.08 & 55.79 & \secondbest{54.08} & \secondbest{54.94} \\
RepNoise                     & \secondbest{10.16} & \secondbest{13.02} & \best{9.63} & \secondbest{54.51} & \best{52.36} & 56.08 \\
BECON                        & \best{1.14} & \best{1.00} & \secondbest{34.33} & \best{47.93} & 54.79 & \best{54.51} \\
\bottomrule
\end{tabular}
\end{adjustbox}
\end{table}

\textbf{Attack robustness against different learning rates and number of epochs.}
Across varied attack hyper-parameters in \cref{tab:experiment_varying_epochs,tab:experiment_varying_learning_rate}, BECON shows a clear immunization effect against harmful-only fine-tuning. Harmful Score remains very low under small learning rates and limited epochs, and BECON continues to resist many stronger attacks with 75-100 epochs or learning rates of \(10^{-4}\) to \(10^{-3}\), where TAR and RepNoise degrade more substantially. At the same time, the increase in HS under aggressive settings such as learning rate of \(5\times10^{-5}\) or epochs of 20 shows that open-weight immunization remains sensitive to the attack budget, and current defenses do not yet provide uniform protection across all fine-tuning configurations. Although RepNoise is stronger in some mid-range settings, BECON offers the most robust overall defense profile. This sensitivity is expected from \cref{thm:main}, where the convergence gap shrinks with more steps and a larger learning rate. BECON therefore slows harmful fine-tuning rather than preventing it. TAR and RepNoise reach a similar harmful score under the strongest attacks, which suggests a limit shared by current open-weight defenses~\citep{rosati2026limits}. This highlights robust protection under larger attack budgets as an important direction for future work.
\begin{table}[htbp]
\caption{\small{Comparison of the utility and fine-tuning capabilities of the non-immunized model and model immunized with BECON.}}
\vspace{2mm}
\label{tab:trainability_and_utility}
\centering
\resizebox{0.96\columnwidth}{!}{
\begin{tabular}{c|cccc|ccc}
\toprule
\multirow{2}{*}{\textbf{Model}} & \multicolumn
{4}{c|}{\textbf{Acc} (\%)} & \multicolumn{3}{c}{\textbf{FA} (\%)} \\
\cmidrule(lr){2-5} \cmidrule(lr){6-8}
& MMLU & TruthfulQA & ARC & Hellaswag & SST2 & AGNEWS & GSM8K \\
\midrule
Non-immunized model & 45.66 & 38.94 & 76.30 & 58.96 & 93.90 & 84.10 & 18.17  \\
\rowcolor{table_color} \textbf{BECON} & 45.14 & 42.54 & 71.74 & 55.26 & 93.50  & 88.47 & 15.97  \\
\bottomrule
\end{tabular}
}
\end{table}

\textbf{Effect of BECON on model utility and trainability.}
Because most users care about direct use and benign fine-tuning, we assess BECON's effects on general capabilities and trainability relative to the base model. \cref{tab:trainability_and_utility} shows that BECON preserves useful capability in several settings, but with a clear trade-off between stronger resistance to harmful fine-tuning and overall performance. It improves TruthfulQA from 38.94\% to 42.54\% but shows visible drops on ARC and other benchmarks. For benign trainability, the immunized model remains learnable on harmless data, matching SST2 and improving AGNEWS from 84.10\% to 88.47\%, though lower GSM8K results show trainability is not fully preserved across tasks. Overall, BECON largely maintains benign usefulness while substantially improving immunization, though remaining utility regressions leave room for improvement.

\subsection{Robust Unlearning against Relearning Attack}
\label{sec:robust_unlearning}
\begin{figure}[t]
\centering
\includegraphics[width=0.58\textwidth]{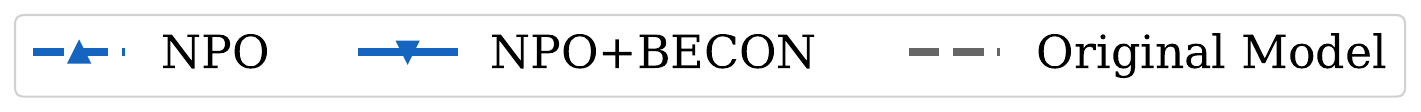}\\
\vspace{1mm}
\includegraphics[width=0.40\textwidth]{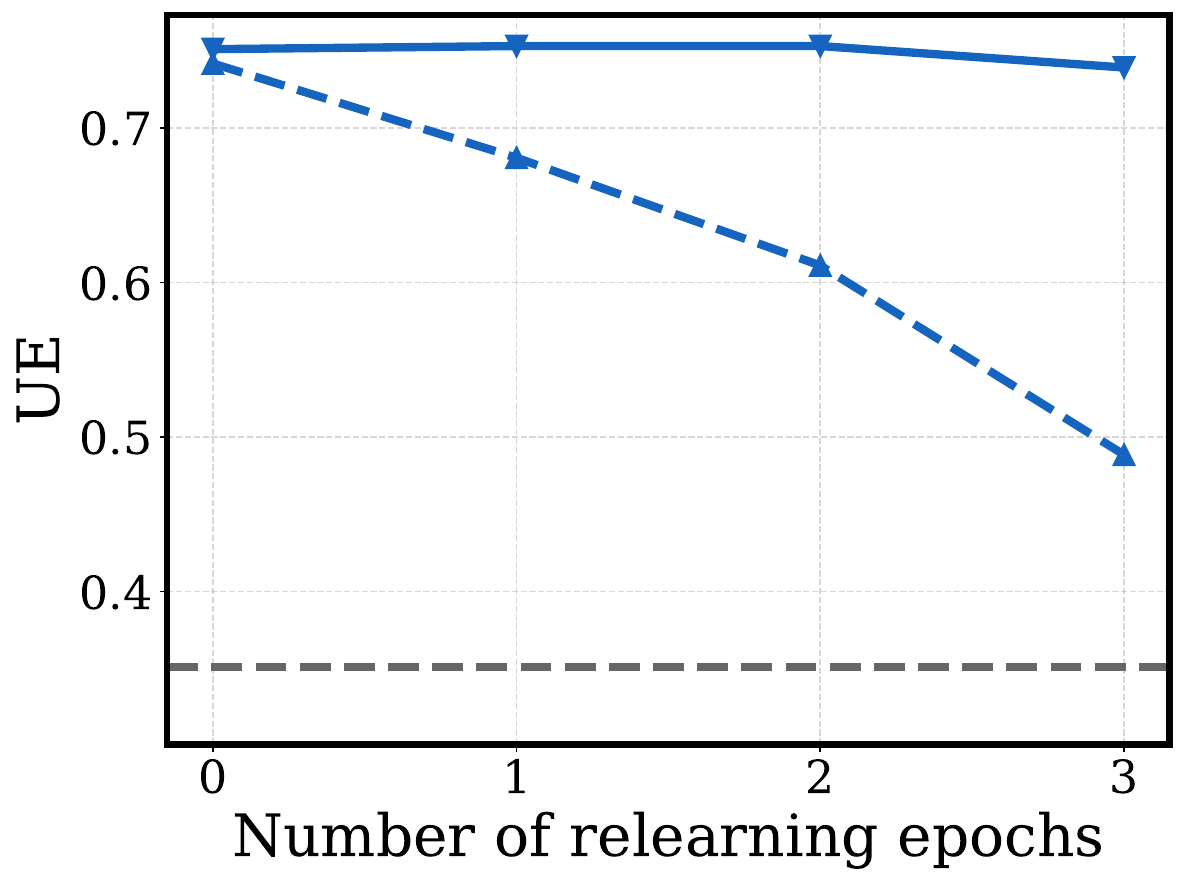}
\hspace{0.01\textwidth}
\includegraphics[width=0.40\textwidth]{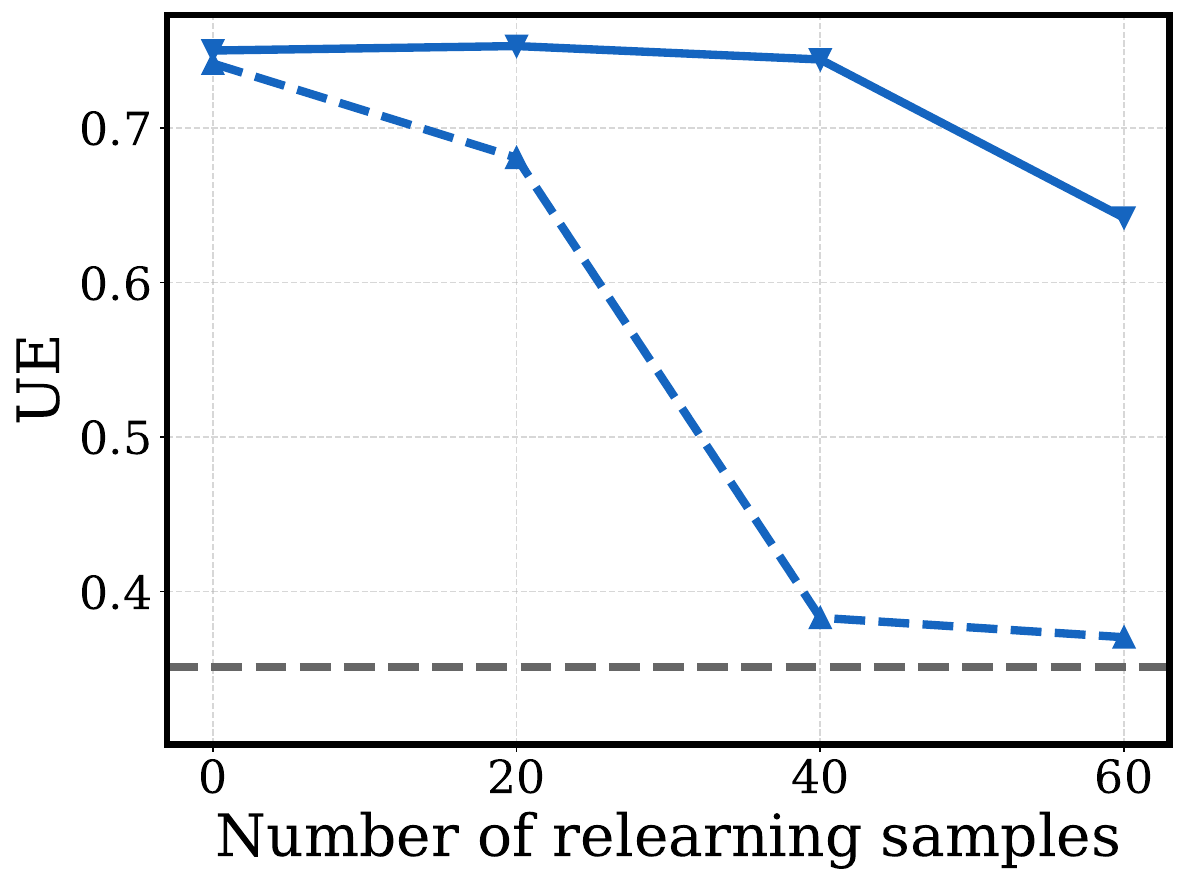}
\caption{Robustness evaluation of unlearning methods on WMDP against relearning attacks. Left: UE vs.\ relearning epochs. Right: UE vs.\ number of relearning samples.}
\label{fig:relearn_wmdp_method}
\vspace*{-2mm}
\end{figure}
In this section, we evaluate the proposed immunization framework for robust unlearning against relearning attacks on Zephyr-7B-beta \citep{tunstall2024zephyr}, focusing on the biosecurity domain of the Weapons of Mass Destruction Proxies benchmark (WMDP) \citep{li2024the}. We compare Negative Preference Optimization (NPO) \citep{zhang2024negative} with NPO+BECON, which additionally includes the $\mathcal{L}_{\text{plateau}}$ and $\mathcal{L}_{\text{cn-min}} - \mathcal{L}_{\text{cn-max}}$ terms from \cref{sect:method}. NPO steers the model away from producing preferred outputs on the forget set by optimizing a loss that assigns negative preference to those targets, thereby discouraging retention of the specified knowledge. Concretely, we use WMDP-Bio as the forget dataset for the unlearning loss and a retain set of unrelated general content from Wikitext \citep{merity2017pointer}, then simulate adversarial relearning by fine-tuning on samples from the WMDP-Bio forget set to test resistance to knowledge recovery. The unlearning robustness performance is quantified using Unlearning Effectiveness (UE) (higher is better), defined as: $\mathrm{UE} = 1 - \texttt{Accuracy on WMDP-Bio evaluation set}$. Further experiment details are provided in Appendix \ref{app:exp_details_unlearning}.

\textbf{Unlearning robustness against relearning attack.} \cref{fig:relearn_wmdp_method} shows that the baseline NPO is highly vulnerable to knowledge recovery. Its UE degrades significantly from $0.74$ to $0.49$ after 3 epochs of relearning. Similarly, increasing the number of relearning samples to 60 causes NPO's UE to decrease to $0.37$. In contrast, NPO+BECON demonstrates remarkable resilience. Under the same conditions, it maintains a high UE of $0.74$ after 3 epochs, showing negligible degradation compared to its initial performance. Even with 60 relearning samples, NPO+BECON retains a substantially higher UE of $0.64$ compared to NPO. These results confirm that our proposed immunization framework effectively prevents the knowledge recovery observed in standard unlearning methods.

\section{Ablation Study}
\begin{table}[!h]
\vspace{-4mm}
\definecolor{darkgreen}{rgb}{0.13, 0.55, 0.13}
\centering
\caption{Ablation of the proposed components on GSM8K. The intermediate rows add each component separately to SFT. BECON combines both. Colored values in parentheses indicate changes from the SFT baseline.}
\label{tab:ablation_proposed_loss}
\resizebox{0.55\textwidth}{!}{
\begin{tabular}{lll}
\toprule
Method & GSM8K HS $\downarrow$ & GSM8K FA $\uparrow$ \\
\midrule
SFT & $23.94$ & $10.90$ \\
SFT + $\mathcal{L}_{\text{cn-min}} - \mathcal{L}_{\text{cn-max}}$ & $20.13$ {\scriptsize\textcolor{darkgreen}{($-3.81$)}} & $15.00$ {\scriptsize\textcolor{darkgreen}{($+4.10$)}} \\
SFT + $\mathcal{L}_{\text{plateau}}$ & $10.40$ {\scriptsize\textcolor{darkgreen}{($-13.54$)}} & $15.07$ {\scriptsize\textcolor{darkgreen}{($+4.17$)}} \\
\rowcolor{table_color}
\textbf{BECON} & \phantom{0}$3.72$ {\scriptsize\textcolor{darkgreen}{($-20.22$)}} & $15.43$ {\scriptsize\textcolor{darkgreen}{($+4.53$)}} \\
\bottomrule
\end{tabular}
}
\end{table}

\textbf{Ablation of Our Components.} Using the experiment setup in \cref{sec:ftaas}, we ablate the individual and combined effects of condition number enlargement and harmful plateau promotion on GSM8K. As shown in Table \ref{tab:ablation_proposed_loss}, condition number enlargement ($\mathcal{L}_{\text{cn-min}} - \mathcal{L}_{\text{cn-max}}$) alone reduces HS to $20.13$. The harmful plateau promotion ($\mathcal{L}_{\text{plateau}}$) is more aggressive, lowering HS to $10.40$ and supporting our theory that condition number manipulation should be complemented by gradient suppression. Combining both components in BECON achieves the best result, reducing HS to $3.72$ while maintaining a high FA of $15.43$, showing that their synergy is crucial for robust immunization. Further discussion on the contribution and effectiveness of the condition-number regularization term is provided in Appendix \ref{app:condition_number_discussion}.

\newpage
\section{Conclusion}
\label{sect:conclusion}
In this paper, we introduce an immunization framework designed to protect LLMs from safety compromise during harmful fine-tuning. By refining the theoretical framework of condition-number-based immunization to prioritize plateau regions and leveraging an efficient estimator for the Hessian, we provide an effective algorithm for LLM immunization. Our evaluations across deployment scenarios, including FTaaS, open-weight settings, and robust unlearning, demonstrate that our method effectively inhibits the optimization of harmful objectives while largely maintaining general model utility and benign trainability. As the demand for open-weight models grows, this efficient approach offers a vital step toward safeguarding safety alignment against adversarial adaptation in real-world environments.

\section{Limitations and Future Work}
\label{app:Limitations_and_future_work}

Our method improves the resilience of open-weight models against harmful fine-tuning, but it does not provide a universal guarantee under all harmful distributions or attack settings. In particular, its protection may weaken when harmful behaviors are substantially different from those represented in the proxy data used during immunization, and stronger or more aggressive fine-tuning can still reduce safety alignment. Moreover, stronger immunization may introduce trade-offs with general utility on some benchmarks. Immunization also adds a one-time training cost, and our analysis assumes deterministic full-batch gradient descent, so extending the bounds to stochastic gradient descent is left to future work. We therefore view this work as a meaningful step toward making harmful adaptation more difficult in practice, while recognizing that an important direction for future work is to better understand and improve robustness under broader, more out-of-distribution, and more resource-intensive attacks while reducing the associated utility trade-offs.

\begin{ack}
Trung Le and Mehrtash Harandi were supported by the ARC Discovery Project grant DP250100262. The authors declare no competing interests.
\end{ack}

\bibliographystyle{unsrtnat}
\bibliography{example_paper}

\appendix
\clearpage

\section{Algorithm}

\begin{algorithm}
\caption{BECON}
\begin{algorithmic}[1]
  \STATE \textbf{Input:} parameters $\bm{\theta}$, plateau radius $\rho$, finite-difference strength $\varepsilon$, number of sub-minibatches $K$, weights $\lambda_{1},\lambda_{2},\lambda_{3}$, optimizer step size $\gamma$
\FOR{each training minibatch}
    \STATE Compute alignment gradient $\bm{g}_{\text{align}} \leftarrow \nabla_{\bm{\theta}}\mathcal{L}_{\text{align}}(\bm{\theta})$.
    \STATE Compute harmful minibatch gradient $\bm{g}_{\mathrm{H}} \leftarrow \nabla_{\bm{\theta}}\mathcal{L}_{\mathrm{H}}(\bm{\theta})$.
    \STATE Plateau perturbation direction (stop-grad):
    \STATE \hspace{1em} $\bm{v} \leftarrow \bm{g}_{\mathrm{H}}/\|\bm{g}_{\mathrm{H}}\|_2$ \;\;(\text{treat } $\bm{v}$ as stop-gradient)
    \STATE \hspace{1em} $\bm{\theta}_{\text{pert}} \leftarrow \bm{\theta} - \rho\,\bm{v}$
    \STATE Plateau gradient estimate:
    \STATE \hspace{1em} $\nabla \mathcal{L}_{\text{plateau}} \leftarrow \bm{g}_{\mathrm{H}} - \nabla_{\bm{\theta}}\mathcal{L}_{\mathrm{H}}(\bm{\theta}_{\text{pert}})$
    \STATE Split the harmful minibatch into $K$ sub-minibatches, each containing $M$ samples.
    \FOR{$k = 1,\dots,K$}
        \STATE Compute sub-minibatch gradient $\bm{g}_k \leftarrow \nabla_{\bm{\theta}}\mathcal{L}_k(\bm{\theta})$ \;\;(\emph{e.g.}, with labels sampled from $p_{\bm{\theta}}(\cdot|\bm{x})$)
        \STATE Compute squared norm $s_k \leftarrow \|\bm{g}_k\|_2^2$
    \ENDFOR
    \STATE Select indices $k_{\max} \leftarrow \arg\max_k s_k$ and $k_{\min} \leftarrow \arg\min_k s_k$
    \STATE Define proxies: $\mathcal{L}_{\text{cn-max}}(\bm{\theta}) \leftarrow s_{k_{\max}}$,\;\; $\mathcal{L}_{\text{cn-min}}(\bm{\theta}) \leftarrow s_{k_{\min}}$
    \STATE Perturb parameters for finite differences:
    \STATE \hspace{1em} $\bm{\theta}'_{\max} \leftarrow \bm{\theta} - \varepsilon \bm{g}_{k_{\max}}$, \;\; $\bm{\theta}'_{\min} \leftarrow \bm{\theta} - \varepsilon \bm{g}_{k_{\min}}$ \;\;(\text{treat } $\bm{g}_{k_{\max}},\bm{g}_{k_{\min}}$ as stop-gradient)
    \STATE Compute gradient differences:
    \STATE \hspace{1em} $d_{\max} \leftarrow \bigl(\nabla_{\bm{\theta}} \mathcal{L}_{k_{\max}}(\bm{\theta}) - \nabla_{\bm{\theta}} \mathcal{L}_{k_{\max}}(\bm{\theta}'_{\max})\bigr) / \varepsilon$
    \STATE \hspace{1em} $d_{\min} \leftarrow \bigl(\nabla_{\bm{\theta}} \mathcal{L}_{k_{\min}}(\bm{\theta}) - \nabla_{\bm{\theta}} \mathcal{L}_{k_{\min}}(\bm{\theta}'_{\min})\bigr) / \varepsilon$
    \STATE Estimate gradients of the proxies:
    \STATE \hspace{1em} $\nabla \mathcal{L}_{\text{cn-max}} \leftarrow 2 d_{\max}$,\;\; $\nabla \mathcal{L}_{\text{cn-min}} \leftarrow 2 d_{\min}$
    \STATE Direct update direction:
    \STATE \hspace{1em} $u \leftarrow \bm{g}_{\text{align}} + \lambda_1 \nabla \mathcal{L}_{\text{plateau}} + \lambda_2 \nabla \mathcal{L}_{\text{cn-min}} - \lambda_3 \nabla \mathcal{L}_{\text{cn-max}}$
    \STATE Update parameters: $\bm{\theta} \leftarrow \bm{\theta} - \gamma \, u$
\ENDFOR
\end{algorithmic}
\label{alg:algorithm}
\end{algorithm}
For the pseudo-label used in Algorithm~\ref{alg:algorithm}, our derivation follows Sophia \citep{liu2024sophia}, where sampling labels from the model predictive distribution gives a theoretical interpretation of the Gauss-Newton term and shows that predictive-distribution gradients provide mathematically grounded probes of local curvature. In implementation, however, we use the ground-truth label as a practical deterministic approximation, rather than sampling a label sequence at each token, since this substantially simplifies computation while maintaining strong empirical performance. This deterministic choice is also adopted in other Sophia-inspired work \citep{zhao-etal-2025-helene}.

\textbf{Regularity assumptions.} The objective in \eqref{eq:overall_formulation} assumes that the harmful loss is twice continuously differentiable near the perturbed parameters, with a bounded and locally Lipschitz Hessian so that the Taylor and finite-difference approximations are accurate. The minimum and maximum over the $K$ sub-minibatches are differentiable when the selected index is unique.
\section{All Proofs}
\label{app:all_proof}

\subsection{Proof of Lemma 3.1}

\noindent\textbf{Lemma 3.1.}

(i) The PL-type inequality holds for $\sigma_{\min}$:
\begin{equation}
\frac{1}{2}\Vert\nabla\mathcal{L}\left(\bm{\theta}\right)\Vert^{2}\geq\sigma_{\min}\left(\mathcal{L}\left(\bm{\theta}\right)-\mathcal{L}^{*}\right), \quad \forall\, \bm{\theta},
\end{equation}
where $\mathcal{L}^{*}=\min_{\bm{\theta}}\mathcal{L}\left(\bm{\theta}\right)$.

(ii) The opposite PL-type inequality holds for $\sigma_{\max}$:
\begin{equation}
\frac{1}{2}\Vert\nabla\mathcal{L}\left(\bm{\theta}\right)\Vert^{2}\leq\sigma_{\max}\left(\mathcal{L}\left(\bm{\theta}\right)-\mathcal{L}^{*}\right), \quad \forall\, \bm{\theta}.
\end{equation}

\textit{Proof of Lemma 3.1}

(i) Using the strongly convex w.r.t. $\sigma_{\min}$, we have
\begin{align*}
\mathcal{L}\left(\bm{\theta}^{*}\right) & \geq\mathcal{L}\left(\bm{\theta}\right)+\left\langle \nabla\mathcal{L}\left(\bm{\theta}\right),\bm{\theta}^{*}-\bm{\theta}\right\rangle +\frac{1}{2}\sigma_{\min}\Vert\bm{\theta}^{*}-\bm{\theta}\Vert^{2}.
\end{align*}
Rearranging the terms, we reach
\begin{align}
\mathcal{L}\left(\bm{\theta}\right)-\mathcal{L}^{*} & \leq\left\langle \nabla\mathcal{L}\left(\bm{\theta}\right),\bm{\theta}-\bm{\theta}^{*}\right\rangle -\frac{1}{2}\sigma_{\min}\Vert\bm{\theta}^{*}-\bm{\theta}\Vert^{2}\nonumber \\
 & \leq\Vert\nabla\mathcal{L}\left(\bm{\theta}\right)\Vert\Vert\bm{\theta}^{*}-\bm{\theta}\Vert-\frac{1}{2}\sigma_{\min}\Vert\bm{\theta}^{*}-\bm{\theta}\Vert^{2}.\label{eq:upper_L_gap_PL}
\end{align}
Denote $r=\Vert\bm{\theta}^{*}-\bm{\theta}\Vert$, we investigate the quadratic function
\[
f\left(r\right)=\Vert\nabla\mathcal{L}\left(\bm{\theta}\right)\Vert r-\frac{1}{2}\sigma_{\min}r^{2},
\]
which is maximized at
\[
r_{\max}=\frac{\Vert\nabla\mathcal{L}\left(\bm{\theta}\right)\Vert}{\sigma_{\min}}.
\]
Substituting $r_{\max}$ back into $f(r)$ yields the maximum value
\[
f_{\max}=\frac{1}{2}\frac{\Vert\nabla\mathcal{L}\left(\bm{\theta}\right)\Vert^{2}}{\sigma_{\min}}.
\]
Linking to (\ref{eq:upper_L_gap_PL}), we reach the conclusion:
\[
\mathcal{L}\left(\bm{\theta}\right)-\mathcal{L}^{*}\leq\frac{1}{2}\frac{\Vert\nabla\mathcal{L}\left(\bm{\theta}\right)\Vert^{2}}{\sigma_{\min}}.
\]
(ii) Using the strongly smooth w.r.t. $\sigma_{\max}$, we obtain for any $\bm{\theta}'$:
\[
\mathcal{L}\left(\bm{\theta}^{'}\right)\leq\mathcal{L}\left(\bm{\theta}\right)+\left\langle \nabla\mathcal{L}\left(\bm{\theta}\right),\bm{\theta}^{'}-\bm{\theta}\right\rangle +\frac{1}{2}\sigma_{\max}\Vert\bm{\theta}^{'}-\bm{\theta}\Vert^{2}.
\]
Choose $\bm{\theta}^{'}=\bm{\theta}-\frac{1}{\sigma_{\max}}\nabla\mathcal{L}\left(\bm{\theta}\right)$ and note that $\mathcal{L}^{*}\leq\mathcal{L}\left(\bm{\theta}^{'}\right)$, we gain

\begin{align*}
\mathcal{L}^{*} & \leq\mathcal{L}\left(\bm{\theta}\right)+\left\langle \nabla\mathcal{L}\left(\bm{\theta}\right),-\frac{1}{\sigma_{\max}}\nabla\mathcal{L}\left(\bm{\theta}\right)\right\rangle +\frac{1}{2}\sigma_{\max}\Vert-\frac{1}{\sigma_{\max}}\nabla\mathcal{L}\left(\bm{\theta}\right)\Vert^{2}\\
 & \leq\mathcal{L}\left(\bm{\theta}\right)-\frac{1}{2\sigma_{\max}}\Vert\nabla\mathcal{L}\left(\bm{\theta}\right)\Vert^{2},
\end{align*}
which is equivalent to
\[
\frac{1}{2}\Vert\nabla\mathcal{L}\left(\bm{\theta}\right)\Vert^{2}\leq\sigma_{\max}\left(\mathcal{L}\left(\bm{\theta}\right)-\mathcal{L}^{*}\right).
\]

\subsection{Proof of Lemma 3.2}

\noindent\textbf{Lemma 3.2.} The update $\bm{\theta}_{t+1}=\bm{\theta}_{t}-\eta\nabla\mathcal{L}\left(\bm{\theta}_{t}\right)$ is efficient in reducing the new objective $\mathcal{L}(\bm{\theta}_{t+1})$ (i.e., $\mathcal{L}\left(\bm{\theta}_{t+1}\right)<\mathcal{L}\left(\bm{\theta}_{t}\right)$) if $0< \eta < \frac{2}{\sigma_{\max}}$.

\textit{Proof of Lemma 3.2.}
Let $h=\Vert\bm{\theta}_{t+1}-\bm{\theta}_{t}\Vert$.
By the second-order Taylor expansion:
\begin{align}
\mathcal{L}\left(\bm{\theta}_{t+1}\right)
 & =\mathcal{L}\left(\bm{\theta}_{t}\right)+\left\langle \bm{\theta}_{t+1}-\bm{\theta}_{t},\nabla\mathcal{L}\left(\bm{\theta}_{t}\right)\right\rangle +\frac{1}{2}\left(\bm{\theta}_{t+1}-\bm{\theta}_{t}\right)^{T}\bm{H}\left(\bm{\theta}_{t+1}-\bm{\theta}_{t}\right)+o(h^{2})\nonumber \\
 & =\mathcal{L}\left(\bm{\theta}_{t}\right)+\left\langle -\eta\nabla\mathcal{L}\left(\bm{\theta}_{t}\right),\nabla\mathcal{L}\left(\bm{\theta}_{t}\right)\right\rangle +\frac{\eta^{2}}{2}\nabla\mathcal{L}\left(\bm{\theta}_{t}\right)^{T}\bm{H}\,\nabla\mathcal{L}\left(\bm{\theta}_{t}\right)+o(h^{2})\nonumber \\
 & \leq\mathcal{L}\left(\bm{\theta}_{t}\right)-\eta\Vert\nabla\mathcal{L}\left(\bm{\theta}_{t}\right)\Vert^{2}+\frac{\sigma_{\max}\eta^{2}}{2}\Vert\nabla\mathcal{L}\left(\bm{\theta}_{t}\right)\Vert^{2}+\epsilon h^{2}\nonumber \\
 & =\mathcal{L}\left(\bm{\theta}_{t}\right)-\eta\Vert\nabla\mathcal{L}\left(\bm{\theta}_{t}\right)\Vert^{2}+\frac{\sigma_{\max}\eta^{2}}{2}\Vert\nabla\mathcal{L}\left(\bm{\theta}_{t}\right)\Vert^{2}+\epsilon\eta^{2}\Vert\nabla\mathcal{L}\left(\bm{\theta}_{t}\right)\Vert^{2}\nonumber \\
 & =\mathcal{L}\left(\bm{\theta}_{t}\right)-\left(1-\frac{\eta}{2}\sigma_{\max}-\epsilon\eta\right)\eta\Vert\nabla\mathcal{L}\left(\bm{\theta}_{t}\right)\Vert^{2},
\label{eq:proof_lemma3.2}
\end{align}
where the small-$o$ notation is defined as $\lim_{h \to 0} o(h^2)/h^2 = 0$. Therefore, in the third line of (\ref{eq:proof_lemma3.2}), for every $\epsilon > 0$, there exists $\delta$ such that whenever $0 < h < \delta$, we have $o(h^2) \leq \epsilon h^2$.

This leads to the conclusion.

\subsection{Proof of Theorem 3.3}

\noindent\textbf{Theorem 3.3.} Considering the learning rate $\eta \in \left(0, \frac{2}{\sigma_{\max}}\right)$, we have the following inequalities:

(i) We can upper-bound $\mathcal{L}(\bm{\theta}_{t+1}) - \mathcal{L}^*$ by
\begin{equation}
\mathcal{L}\left(\bm{\theta}_{t+1}\right)-\mathcal{L}^{*}\leq\left(1-\frac{1}{\kappa}\right)^{t+1}\left(\mathcal{L}\left(\bm{\theta}_{0}\right)-\mathcal{L}^{*}\right)+\frac{\left(\sigma_{\max}\eta-1\right)^{2}}{2\sigma_{\max}}\sum_{i=1}^{t}\left(1-\frac{1}{\kappa}\right)^{t-i}\Vert\nabla\mathcal{L}\left(\bm{\theta}_{i}\right)\Vert^{2}.
\end{equation}
(ii) We can lower-bound $\mathcal{L}(\bm{\theta}_{t+1}) - \mathcal{L}^*$ by 

$\square$ If $\kappa > 2$ and $\eta\in\left[\frac{\sigma_{\min}}{\sigma_{\max}^{2}},\frac{2}{\sigma_{\max}}\right)$, we have the lower bound
\begin{equation}
\begin{aligned}
\mathcal{L}\left(\bm{\theta}_{t+1}\right)-\mathcal{L}^{*}\geq & \left(1-\frac{1}{\kappa}\right)^{3\left(t+1\right)}\left(\mathcal{L}\left(\bm{\theta}_{0}\right)-\mathcal{L}^{*}\right)- \\
& \left(\eta-\frac{\sigma_{\min}}{\sigma_{\max}^{2}}\right)\left(1-\frac{1}{\kappa^{2}}\right)\sum_{i=1}^{t}\left(1-\frac{1}{\kappa}\right)^{3\left(t-i\right)}\Vert\nabla\mathcal{L}\left(\bm{\theta}_{i}\right)\Vert^{2}.
\end{aligned}
\end{equation}
$\square$ If the learning rate $0<\eta<\frac{\sigma_{\min}}{\sigma_{\max}^{2}}$, we have the lower bound
\begin{align}
\mathcal{L}\left(\bm{\theta}_{t+1}\right)-\mathcal{L}^{*} & \geq\left(1-\frac{1}{\kappa}\right)^{t+1}\left(\mathcal{L}\left(\bm{\theta}_{0}\right)-\mathcal{L}^{*}\right)-\nonumber \\
\frac{\eta}{2}\left(1-\frac{\sigma_{\min}\eta}{2}\right) & \sum_{i=1}^{t}\left(1-\frac{1}{\kappa}\right)^{t-i}\Vert\nabla\mathcal{L}\left(\bm{\theta}_{i}\right)\Vert^{2}.
\end{align}

\textit{Proof of Theorem 3.3.}

(i) Using the second-order Taylor expansion, we have 
\begin{align*}
\mathcal{L}\left(\bm{\theta}_{t+1}\right) & \leq\mathcal{L}\left(\bm{\theta}_{t}\right)-\left(1-\frac{\sigma_{\max}\eta}{2}\right)\eta\Vert\nabla\mathcal{L}\left(\bm{\theta}_{t}\right)\Vert^{2}\\
 & =\mathcal{L}\left(\bm{\theta}_{t}\right)-\frac{1}{2\sigma_{\max}}\Vert\nabla\mathcal{L}\left(\bm{\theta}_{t}\right)\Vert^{2}+\left(\frac{1}{2\sigma_{\max}}-\eta+\frac{\sigma_{\max}}{2}\eta^{2}\right)\Vert\nabla\mathcal{L}\left(\bm{\theta}_{t}\right)\Vert^{2}\\
 & =\mathcal{L}\left(\bm{\theta}_{t}\right)-\frac{1}{2\sigma_{\max}}\Vert\nabla\mathcal{L}\left(\bm{\theta}_{t}\right)\Vert^{2}+\frac{\left(\sigma_{\max}\eta-1\right)^{2}}{2\sigma_{\max}}\Vert\nabla\mathcal{L}\left(\bm{\theta}_{t}\right)\Vert^{2}.
\end{align*}
Deducting $\mathcal{L}^*$ from two sides and using PL-typed inequality, we obtain
\begin{align}
\mathcal{L}\left(\bm{\theta}_{t+1}\right)-\mathcal{L}^{*} & \leq\mathcal{L}\left(\bm{\theta}_{t}\right)-\mathcal{L}^{*}-\frac{1}{2\sigma_{\max}}\Vert\nabla\mathcal{L}\left(\bm{\theta}_{t}\right)\Vert^{2}+\frac{\left(\sigma_{\max}\eta-1\right)^{2}}{2\sigma_{\max}}\Vert\nabla\mathcal{L}\left(\bm{\theta}_{t}\right)\Vert^{2}\nonumber \\
\leq & \mathcal{L}\left(\bm{\theta}_{t}\right)-\mathcal{L}^{*}-\frac{\sigma_{\min}}{\sigma_{\max}}\left(\mathcal{L}\left(\bm{\theta}_{t}\right)-\mathcal{L}^{*}\right)+\frac{\left(\sigma_{\max}\eta-1\right)^{2}}{2\sigma_{\max}}\Vert\nabla\mathcal{L}\left(\bm{\theta}_{t}\right)\Vert^{2}\nonumber \\
\leq & \left(1-\frac{\sigma_{\min}}{\sigma_{\max}}\right)\left(\mathcal{L}\left(\bm{\theta}_{t}\right)-\mathcal{L}^{*}\right)+\frac{\left(\sigma_{\max}\eta-1\right)^{2}}{2\sigma_{\max}}\Vert\nabla\mathcal{L}\left(\bm{\theta}_{t}\right)\Vert^{2}.\label{eq:gap}
\end{align}
Applying inequality (\ref{eq:gap}) recursively, we reach
\[
\mathcal{L}\left(\bm{\theta}_{t+1}\right)-\mathcal{L}^{*}\leq\left(1-\frac{\sigma_{\min}}{\sigma_{\max}}\right)^{t+1}\left(\mathcal{L}\left(\bm{\theta}_{0}\right)-\mathcal{L}^{*}\right)+\sum_{i=1}^{t}\left(1-\frac{\sigma_{\min}}{\sigma_{\max}}\right)^{t-i}\frac{\left(\sigma_{\max}\eta-1\right)^{2}}{2\sigma_{\max}}\Vert\nabla\mathcal{L}\left(\bm{\theta}_{i}\right)\Vert^{2}.
\]

(ii) Using the second-order Taylor expansion, we have
\begin{align}
\mathcal{L}\left(\bm{\theta}_{t+1}\right) & \approx\mathcal{L}\left(\bm{\theta}_{t}\right)+\left\langle \bm{\theta}_{t+1}-\bm{\theta}_{t},\nabla\mathcal{L}\left(\bm{\theta}_{t}\right)\right\rangle +\frac{1}{2}\left(\bm{\theta}_{t+1}-\bm{\theta}_{t}\right)^{T}\bm{H}\left(\bm{\theta}_{t+1}-\bm{\theta}_{t}\right)\nonumber \\
 & \geq\mathcal{L}\left(\bm{\theta}_{t}\right)-\eta\Vert\nabla\mathcal{L}\left(\bm{\theta}_{t}\right)\Vert^{2}+\frac{1}{2}\sigma_{\min}\Vert\bm{\theta}_{t+1}-\bm{\theta}_{t}\Vert^{2}\nonumber \\
 & =\mathcal{L}\left(\bm{\theta}_{t}\right)-\eta\Vert\nabla\mathcal{L}\left(\bm{\theta}_{t}\right)\Vert^{2}+\frac{1}{2}\sigma_{\min}\eta^{2}\Vert\nabla\mathcal{L}\left(\bm{\theta}_{t}\right)\Vert^{2}\nonumber \\
 & =\mathcal{L}\left(\bm{\theta}_{t}\right)-\left(1-\frac{\sigma_{\min}\eta}{2}\right)\eta\Vert\nabla\mathcal{L}\left(\bm{\theta}_{t}\right)\Vert^{2}.\label{eq:eta_range_inq-1}
\end{align}
Deducting $\mathcal{L}^*$ to two sides of (\ref{eq:eta_range_inq-1}), we gain
\begin{align*}
\mathcal{L}\left(\bm{\theta}_{t+1}\right)-\mathcal{L}^{*} & \geq\mathcal{L}\left(\bm{\theta}_{t}\right)-\mathcal{L}^{*}-\left(1-\frac{\sigma_{\min}\eta}{2}\right)\eta\Vert\nabla\mathcal{L}\left(\bm{\theta}_{t}\right)\Vert^{2}\\
 & =\mathcal{L}\left(\bm{\theta}_{t}\right)-\mathcal{L}^{*}-h\left(\eta\right)\Vert\nabla\mathcal{L}\left(\bm{\theta}_{t}\right)\Vert^{2},
\end{align*}
where $h\left(\eta\right)=\eta-\frac{\sigma_{\min}}{2}\eta^{2}$.

We consider two following cases:

* \textbf{Case 1}: $\eta \in \left[\frac{\sigma_{\min}}{\sigma_{\max}^{2}}, \frac{2}{\sigma_{\max}} \right)$

We have
\begin{align*}
\mathcal{L}\left(\bm{\theta}_{t+1}\right)-\mathcal{L}^{*} & \ge\mathcal{L}\left(\bm{\theta}_{t}\right)-\mathcal{L}^{*}-h\left(\frac{\sigma_{\min}}{\sigma_{\max}^{2}}\right)\Vert\nabla\mathcal{L}\left(\bm{\theta}_{t}\right)\Vert^{2}+\left(h\left(\frac{\sigma_{\min}}{\sigma_{\max}^{2}}\right)-h\left(\eta\right)\right)\Vert\nabla\mathcal{L}\left(\bm{\theta}_{t}\right)\Vert^{2}\\
= & \mathcal{L}\left(\boldsymbol{\theta}_{t}\right)-\mathcal{L}^{*}-h\left(\frac{\sigma_{\min}}{\sigma_{\max}^{2}}\right)\Vert\nabla\mathcal{L}\left(\boldsymbol{\theta}_{t}\right)\Vert^{2}+\left(\frac{\sigma_{\min}}{\sigma_{\max}^{2}}-\eta\right)\left[1-\frac{\sigma_{\min}}{2}\left(\frac{\sigma_{\min}}{\sigma_{\max}^{2}}+\eta\right)\right]\\
\geq & \mathcal{L}\left(\boldsymbol{\theta}_{t}\right)-\mathcal{L}^{*}-h\left(\frac{\sigma_{\min}}{\sigma_{\max}^{2}}\right)\Vert\nabla\mathcal{L}\left(\boldsymbol{\theta}_{t}\right)\Vert^{2}-\left(\eta-\frac{\sigma_{\min}}{\sigma_{\max}^{2}}\right)\left(1-\left(\frac{\sigma_{\min}}{\sigma_{\max}}\right)^{2}\right)\Vert\nabla\mathcal{L}\left(\boldsymbol{\theta}_{t}\right)\Vert^{2}.
\end{align*}By noting that
\[
\Vert\nabla\mathcal{L}\left(\boldsymbol{\theta}_{t}\right)\Vert^{2}\leq2\sigma_{\max}\left(\mathcal{L}\left(\boldsymbol{\theta}_{t}\right)-\mathcal{L}^{*}\right),
\]
we gain
\begin{align}
\mathcal{L}\left(\boldsymbol{\theta}_{t+1}\right)-\mathcal{L}^{*} & \geq\mathcal{L}\left(\boldsymbol{\theta}_{t}\right)-\mathcal{L}^{*}-2\sigma_{\max}h\left(\frac{\sigma_{\min}}{\sigma_{\max}^{2}}\right)\left(\mathcal{L}\left(\boldsymbol{\theta}_{t}\right)-\mathcal{L}^{*}\right)\nonumber \\
 & -\left(\eta-\frac{\sigma_{\min}}{\sigma_{\max}^{2}}\right)\left(1-\left(\frac{\sigma_{\min}}{\sigma_{\max}}\right)^{2}\right)\Vert\nabla\mathcal{L}\left(\boldsymbol{\theta}_{t}\right)\Vert^{2}\nonumber \\
 & =\mathcal{L}\left(\boldsymbol{\theta}_{t}\right)-\mathcal{L}^{*}-\left(2\frac{\sigma_{\min}}{\sigma_{\max}}-\frac{\sigma_{\min}^{3}}{\sigma_{\max}^{3}}\right)\left(\mathcal{L}\left(\boldsymbol{\theta}_{t}\right)-\mathcal{L}^{*}\right)\nonumber \\
 & -\left(\eta-\frac{\sigma_{\min}}{\sigma_{\max}^{2}}\right)\left(1-\left(\frac{\sigma_{\min}}{\sigma_{\max}}\right)^{2}\right)\Vert\nabla\mathcal{L}\left(\boldsymbol{\theta}_{t}\right)\Vert^{2}\nonumber \\
\geq & \left(1-2\frac{\sigma_{\min}}{\sigma_{\max}}+\frac{\sigma_{\min}^{3}}{\sigma_{\max}^{3}}\right)\left(\mathcal{L}\left(\boldsymbol{\theta}_{t}\right)-\mathcal{L}^{*}\right)-\left(\eta-\frac{\sigma_{\min}}{\sigma_{\max}^{2}}\right)\left(1-\left(\frac{\sigma_{\min}}{\sigma_{\max}}\right)^{2}\right)\Vert\nabla\mathcal{L}\left(\boldsymbol{\theta}_{t}\right)\Vert^{2}\nonumber \\
\stackrel{(1)}{\geq} & \left(1-\frac{\sigma_{\min}}{\sigma_{\max}}\right)^{3}\left(\mathcal{L}\left(\boldsymbol{\theta}_{t}\right)-\mathcal{L}^{*}\right)-\left(\eta-\frac{\sigma_{\min}}{\sigma_{\max}^{2}}\right)\left(1-\left(\frac{\sigma_{\min}}{\sigma_{\max}}\right)^{2}\right)\Vert\nabla\mathcal{L}\left(\boldsymbol{\theta}_{t}\right)\Vert^{2}.\label{eq:case1_Lgap}
\end{align}

The condition $\kappa>2$, equivalently $\alpha=\sigma_{\min}/\sigma_{\max}<1/2$, is used here as a technical assumption in the analysis to show that \textbf{a large condition number alone is insufficient, and that an additional plateau loss term is needed for effective immunization.} We note that we have (1) because if $\alpha = \frac{\sigma_{\min}}{\sigma_{\max}} <1/2$ then
\begin{align*}
1-2\frac{\sigma_{\min}}{\sigma_{\max}}+\frac{\sigma_{\min}^{3}}{\sigma_{\max}^{3}} & =1-2\alpha+\alpha^{3}=\left(1-\alpha\right)^{3}+\alpha-3\alpha^{2}+2\alpha^{3}\\
= & \left(1-\alpha\right)^{3}+\alpha\left(1-\alpha\right)\left(1-2\alpha\right)>\left(1-\alpha\right)^{3}.
\end{align*}
Applying (\ref{eq:case1_Lgap}) recursively, we obtain
\[
\begin{aligned}
\mathcal{L}\left(\boldsymbol{\theta}_{t+1}\right)-\mathcal{L}^{*}\geq &
\left(1-\frac{\sigma_{\min}}{\sigma_{\max}}\right)^{3\left(t+1\right)}\left(\mathcal{L}\left(\boldsymbol{\theta}_{0}\right)-\mathcal{L}^{*}\right) {}- \\
& \left(\eta-\frac{\sigma_{\min}}{\sigma_{\max}^{2}}\right)\left(1-\left(\frac{\sigma_{\min}}{\sigma_{\max}}\right)^{2}\right)\sum_{i=1}^{t}\left(1-\frac{\sigma_{\min}}{\sigma_{\max}}\right)^{3\left(t-i\right)}\Vert\nabla\mathcal{L}\left(\boldsymbol{\theta}_{i}\right)\Vert^{2}.
\end{aligned}
\]

* \textbf{Case 2}: $\eta \in \left(0 , \frac{\sigma_{\min}}{\sigma_{\max}^{2}} \right]$

We derive as
\begin{align}
\mathcal{L}\left(\boldsymbol{\theta}_{t+1}\right)-\mathcal{L}^{*} & \stackrel{(1)}{\ge}\mathcal{L}\left(\boldsymbol{\theta}_{t}\right)-\mathcal{L}^{*}-\frac{1}{2}h\left(\frac{\sigma_{\min}}{\sigma_{\max}^{2}}\right)\Vert\nabla\mathcal{L}\left(\boldsymbol{\theta}_{t}\right)\Vert^{2}-\frac{1}{2}h\left(\eta\right)\Vert\nabla\mathcal{L}\left(\boldsymbol{\theta}_{t}\right)\Vert^{2}\nonumber \\
 & \geq\mathcal{L}\left(\boldsymbol{\theta}_{t}\right)-\mathcal{L}^{*}-\sigma_{\max}h\left(\frac{\sigma_{\min}}{\sigma_{\max}^{2}}\right)\left(\mathcal{L}\left(\boldsymbol{\theta}_{t}\right)-\mathcal{L}^{*}\right)-\frac{\eta}{2}\left(1-\frac{\sigma_{\min}\eta}{2}\right)\Vert\nabla\mathcal{L}\left(\boldsymbol{\theta}_{t}\right)\Vert^{2}\nonumber \\
 & \geq\left(1-\frac{\sigma_{\min}}{\sigma_{\max}}+\frac{\sigma_{\min}^{3}}{2\sigma_{\max}^{3}}\right)\left(\mathcal{L}\left(\boldsymbol{\theta}_{t}\right)-\mathcal{L}^{*}\right)-\frac{\eta}{2}\left(1-\frac{\sigma_{\min}\eta}{2}\right)\Vert\nabla\mathcal{L}\left(\boldsymbol{\theta}_{t}\right)\Vert^{2}\nonumber \\
 & \geq\left(1-\frac{\sigma_{\min}}{\sigma_{\max}}\right)\left(\mathcal{L}\left(\boldsymbol{\theta}_{t}\right)-\mathcal{L}^{*}\right)-\frac{\eta}{2}\left(1-\frac{\sigma_{\min}\eta}{2}\right)\Vert\nabla\mathcal{L}\left(\boldsymbol{\theta}_{t}\right)\Vert^{2}.\label{eq:case2_gapL}
\end{align}
Applying (\ref{eq:case2_gapL}) recursively, we gain
\[
\begin{aligned}
\mathcal{L}\left(\boldsymbol{\theta}_{t+1}\right)-\mathcal{L}^{*}\geq &
\left(1-\frac{\sigma_{\min}}{\sigma_{\max}}\right)^{t+1}\left(\mathcal{L}\left(\boldsymbol{\theta}_{0}\right)-\mathcal{L}^{*}\right) {}- \\
& \frac{\eta}{2}\left(1-\frac{\sigma_{\min}\eta}{2}\right)\sum_{i=1}^{t}\left(1-\frac{\sigma_{\min}}{\sigma_{\max}}\right)^{t-i}\Vert\nabla\mathcal{L}\left(\boldsymbol{\theta}_{i}\right)\Vert^{2}.
\end{aligned}
\]

\subsection{Proof of Theorem 4.1.}
\label{sec:hessian_matrix_estimation}

\noindent\textbf{Theorem 4.1.} Consider a harmful minibatch of size $B$ partitioned into $K$ sub-minibatches of size $M$ ($B=KM$).
Let $\bm{g}_k$ be the average gradient of the $k$-th sub-minibatch computed with labels sampled from the model's predictive distribution.
Define $\bm{G}=[\bm{g}_1,\dots,\bm{g}_K]\in\mathbb{R}^{d\times K}$. The Gram matrix estimator of the minibatch Gauss-Newton curvature is given by:
\begin{equation}
\widehat{\bm{H}}_{\mathrm{GN}} \coloneqq \frac{M^2}{B} \bm{G} \bm{G}^\top.
\end{equation}

\noindent\textit{Proof of Theorem~\ref{thm:hessian_estimation_ce}.} Let $\bm{f}(\bm{\theta},\bm{x})\in\mathbb{R}^{V}$ be the logits and $p_{\bm{\theta}}(\cdot\mid \bm{x})\coloneqq \mathrm{softmax}(\bm{f}(\bm{\theta},\bm{x}))$.
For cross-entropy loss $\ell(\cdot,\cdot)$, a key identity (see \citep{liu2024sophia}) expresses the Gauss-Newton term as a gradient outer-product expectation:
\begin{equation}
\bm{H}_{\mathrm{GN}}(\bm{x};\bm{\theta})
\;\coloneqq\;
\bm{J}_{\bm{\theta}} \bm{f}(\bm{\theta}, \bm{x}) \cdot \bm{S} \cdot \bm{J}_{\bm{\theta}} \bm{f}(\bm{\theta}, \bm{x})^{\top}
\;=\;
\mathbb{E}_{\bm{\hat y}\sim p_{\bm{\theta}}(\cdot\mid \bm{\hat{x}})}
\Big[
  \nabla_{\bm{\theta}} \ell\bigl(\bm{f}(\bm{\theta},\bm{x}), \bm{\hat y}\bigr)\,
  \nabla_{\bm{\theta}} \ell\bigl(\bm{f}(\bm{\theta},\bm{x}),\bm{\hat y}\bigr)^\top
\Big].
\label{eq:gn_outer_product_identity}
\end{equation}

Consider a harmful minibatch of size $B$ partitioned into $K$ sub-minibatches of size $M$ ($B=KM$).
For each example $\bm{x}_{k,m}$, sample a label $\bm{\hat y}_{k,m}\sim p_{\bm{\theta}}(\cdot\mid \bm{x}_{k,m})$ independently and define the per-example gradient
\[
  \bm{g}_{k,m}\;\coloneqq\;\nabla_\theta \ell\bigl(\bm{f}(\bm{\theta},\bm{x}_{k,m}),\bm{\hat y}_{k,m}\bigr)\in\mathbb{R}^{d}.
\]
Define the sub-minibatch gradients $\bm{g}_k\coloneqq \frac{1}{M}\sum_{m=1}^{M}\bm{g}_{k,m}$ and $\bm{G}=[\bm{g}_1,\dots,\bm{g}_K]\in\mathbb{R}^{d\times K}$.

By \eqref{eq:gn_outer_product_identity}, the minibatch Gauss-Newton curvature is
\[
\bm{H}_{\mathrm{GN}}^{(B)}(\theta)
\;=\;
\frac{1}{B}\sum_{k=1}^{K}\sum_{m=1}^{M}\mathbb{E}\bigl[\bm{g}_{k,m}\bm{g}_{k,m}^\top\bigr],
\]
where the expectation is over the sampled labels $\{\bm{\hat y}_{k,m}\}$ (conditioning on the minibatch inputs).
Moreover, the first Bartlett identity implies $\mathbb{E}[\bm{g}_{k,m}]=0$ for each $(k,m)$, and by independence across examples, for $m\neq m'$,
\[
\mathbb{E}\bigl[\bm{g}_{k,m}\bm{g}_{k,m'}^\top\bigr]
=
\mathbb{E}[\bm{g}_{k,m}]\,\mathbb{E}[\bm{g}_{k,m'}]^\top
=0.
\]
Therefore,
\[
\mathbb{E}\bigl[\bm{g}_k \bm{g}_k^\top\bigr]
\,=\,
\frac{1}{M^2}\sum_{m=1}^{M}\mathbb{E}\bigl[\bm{g}_{k,m}\bm{g}_{k,m}^\top\bigr].
\]
Summing over $k$ and using $\bm{G}\bm{G}^\top=\sum_{k=1}^{K}\bm{g}_k \bm{g}_k^\top$, we have
\[
\mathbb{E}\bigl[\bm{G}\bm{G}^\top\bigr]
\,=\,
\frac{1}{M^2}\sum_{k=1}^{K}\sum_{m=1}^{M}\mathbb{E}\bigl[\bm{g}_{k,m}\bm{g}_{k,m}^\top\bigr].
\]
Multiplying both sides by $\frac{M^2}{B}$ yields
\begin{equation}
\mathbb{E}\!\left[\frac{M^2}{B}\bm{G}\bm{G}^\top\right]
\,=\,
\frac{1}{B}\sum_{k=1}^{K}\sum_{m=1}^{M}\mathbb{E}\bigl[\bm{g}_{k,m}\bm{g}_{k,m}^\top\bigr]
\,=\,
\bm{H}_{\mathrm{GN}}^{(B)}(\bm{\theta}),
\end{equation}
so $\widehat{\bm{H}}_{\mathrm{GN}}\coloneqq \frac{M^2}{B}\bm{G}\bm{G}^\top$ is an unbiased Gram-matrix estimator of the minibatch Gauss-Newton curvature. This proves the theorem.

\section{Experimental Details}
\label{app:experiment_details}

\subsection{Defense against harmful fine-tuning in FTaaS}
\label{app:exp_details}

\subsubsection{Datasets}
\label{app:exp_details_ftaas}

The Stanford Sentiment Treebank (SST2) \citep{socher2013recursive} is a binary classification dataset comprising sentences extracted from movie reviews, each labeled as either positive or negative sentiment. The prompt format for this dataset is as follows:
\begin{tcolorbox}[colback=gray!20!white, colbacktitle=white, coltitle=black, colframe=black!75!black, boxrule=0.7pt, halign title=center, title=\textbf{Prompt format with example input for SST2}]
Below is an instruction that describes a task, paired with an input that provides further context. Write a response that appropriately completes the request.\\ \\
\#\#\# Instruction:\\
Analyze the sentiment of the input, and respond only positive or negative.\\ \\
\#\#\# Input: \\
\textcolor{blue}{a casual intelligence} \\ \\
\#\#\# Response:\\
\end{tcolorbox}

The AGNEWS \citep{zhang2015character} dataset is a collection of news articles categorized into four classes: World, Sports, Business, and Science/Technology. It contains 120k samples in the training split and 7.9k samples in the test split. The prompt format for this dataset is as follows:
\begin{tcolorbox}[colback=gray!20!white, colbacktitle=white, coltitle=black, colframe=black!75!black, boxrule=0.7pt, halign title=center, title=\textbf{Prompt format with example input for AGNEWS}]
Below is an instruction that describes a task, paired with an input that provides further context. Write a response that appropriately completes the request.\\ \\
\#\#\# Instruction:\\
Categorize the news article given in the input into one of the 4 categories:\\ \\World\\Sports\\Business\\Sci/Tech\\ \\ \\
\#\#\# Input: \\
\textcolor{blue}{
Einstein's warp effect measured Two scientists beat a \$600 million Nasa mission to be first to measure a prediction of Einstein's relativity theory.
}\\ \\
\#\#\# Response:\\
\end{tcolorbox}

The (Grade School Math 8K) GSM8K \citep{cobbe2021training} is a question-answering dataset comprising 8.5k grade-school math problems. These problems
typically require 2-8 steps to solve, primarily involving a sequence of elementary calculations using basic arithmetic operations to arrive at the
final answer. The prompt format for this dataset is as follows:
\begin{tcolorbox}[colback=gray!20!white, colbacktitle=white, coltitle=black, colframe=black!75!black, boxrule=0.7pt, halign title=center, title=\textbf{Prompt format with example instruction for GSM8K}]
Below is an instruction that describes a task. Write a response that appropriately completes the request.\\ \\
\#\#\# Instruction:\\
\textcolor{blue}{
Natalia sold clips to 48 of her friends in April, and then she sold half as many clips in May. How many clips did Natalia sell altogether in April and May?}\\ \\
\#\#\# Response:\\
\end{tcolorbox}

The harmful instructions used in our experiments are from the BeaverTails dataset \citep{ji2023beavertails}. This dataset consists of more than 300k pairs of human-labeled question-answer spanning across 14 harm categories, where each response could be flagged with more than one of those categories. Other works \citep{rosati2024representation, huang2025booster} use the harmful dataset from \citep{rosati2024representation}, which is a filtered subset of BeaverTails-30k-train for alignment and fine-tuning attack, and use a filtered subset of BeaverTails-30k-test for evaluation. However, based on our observation and as pointed out by \citep{qi2025on}, these datasets suffer from an overlapped problem between the alignment split, fine-tuning attack split, and evaluation split. Therefore, to ensure a fair evaluation, we construct three separate and non-overlapping data splits for our experiments:

\begin{enumerate}[leftmargin=*]
\item \textbf{Immunization step.} We perform filtering of the dataset in \citep{rosati2024representation} to come up with a subset of 4972 samples for model alignment.

\item \textbf{Fine-tuning attack step.} We use the filtered version of BeaverTails-330k curated by \citep{qi2025on}, including 4986 samples, as the harmful samples that is used to poison the benign dataset.

\item \textbf{Evaluation.} We evaluate the Harmful Score of the fine-tuned model on the filtered subset of the BeaverTails-Evaluation dataset \citep{ji2023beavertails}, which contains 699 malicious questions from 14 categories. We compute Harmful Score using the QA-Moderation model from \citep{ji2023beavertails}, a classifier that judges whether a question-answer pair is safe or unsafe. QA-Moderation is widely used in prior LLM safety work because it has strong agreement with human judgments while requiring no API cost. We do not conduct additional human evaluation. Instead, we rely on the human-feedback agreement reported by \citep{ji2023beavertails}, where QA-Moderation achieves agreement rates ranging from 80\% to 100\% across the tested LLMs.

\end{enumerate}

\subsubsection{Training Details}

Consistent with prior research \citep{huang2024lisa, huang2024vaccine, huang2025booster}, we employ LoRA \citep{hu2022lora} with a rank of 32 and an alpha of 4 to update model parameters during both the alignment and fine-tuning attack phases. All experiments are conducted on a NVIDIA A100, H100 or H200 GPU using a batch size of 16, except for the Gemma2-9B model which requires a batch size of 10 due to memory constraints, and we report the average results across three random seeds. For the immunizaiton step, we train for 20 epochs using the AdamW optimizer \citep{loshchilov2018decoupled} with a learning rate of $5 \times 10^{-4}$ and weight decay of 0.1, while our method's hyperparameters are set to $\rho = \epsilon = 1 \times 10^{-1}$, $\lambda_1 = 5.0$, $\tau=0.5$, and $\lambda_2 = \lambda_3 = 0.1$. The sub-minibatch size is 4. In the subsequent fine-tuning attack phase, we train on the SST2, AGNEWS, and GSM8K datasets for 20 epochs using AdamW, adjusting the learning rate to $1 \times 10^{-5}$ and incorporating a warmup ratio of 0.1 alongside a weight decay of 0.1.

\subsubsection{Metrics and Evaluation}

We evaluate the downstream datasets based on the percentage of exact matches between the generated responses and the ground truth in the test sets:

\begin{itemize}[leftmargin=15pt]
  \item \textbf{SST2}: A prediction is counted as correct if it exactly matches the labels "positive" or "negative".
  \item \textbf{AGNEWS}: A prediction is counted as correct if it aligns with one of the standard category labels (World, Sports, Business, or Sci/Tech).
  \item \textbf{GSM8K}: A prediction is counted as correct if the concluded answer in the response matches the ground truth solution.
\end{itemize}

\subsubsection{Baselines}
This section describes the baseline methods used in our experiments:

\begin{itemize}[leftmargin=15pt]
  \item \textbf{Supervised Fine-Tuning (SFT)} is the standard approach: we fine-tune the model on the alignment dataset during the alignment stage, and fine-tune it on the harmful dataset during the fine-tuning attack stage.
  \item \textbf{Vaccine} \citep{huang2024vaccine} is an alignment-stage defense that improves robustness to harmful fine-tuning by increasing resilience to harmful embedding perturbations. Vaccine also adopts a double-LoRA setup: the LoRA learned during alignment is merged into the base model before the attack stage, and a new LoRA is introduced for the subsequent fine-tuning.
  \item \textbf{Booster} \citep{huang2025booster} is an alignment-stage method that, besides fitting the alignment data, aims to reduce how quickly the harmful loss can be decreased. Concretely, it minimizes the \emph{harmful loss reduction rate}, defined as the gap between the harmful loss of the current model and the harmful loss after one gradient-descent step on harmful data.
  \item \textbf{Lisa} \citep{huang2024lisa} is a fine-tuning-stage defense that mixes alignment data into the fine-tuning process. It alternates between optimizing on the alignment dataset and on the harmful fine-tuning dataset, and uses a proximal regularizer to limit model drift between consecutive states for more stable training.
\end{itemize}

\subsection{Immunization against harmful fine-tuning attacks in open-weight settings}
\label{app:exp_details_immunization}

Here we detail the implementation details of the baseline defense methods and our method:
\begin{itemize}[leftmargin=15pt]
  \item \textbf{TAR} \citep{tamirisa2025tamperresistant}: We use batch size $= 10$. Hyperparameters: $\lambda_{\mathrm{TR}} = 2$ and $\lambda_{\mathrm{retain}} = 1$.
  \item \textbf{RepNoise} \citep{rosati2024representation}: We use batch size $= 10$. Hyperparameters: $\alpha = 1$ and $\beta = 0.001$.
  \item \textbf{BECON}: We use the same hyperparameters as in \cref{app:exp_details_ftaas}.
\end{itemize}

We reuse the same non-overlapping BeaverTails data construction described in \cref{app:exp_details_ftaas} for harmful fine-tuning resistance evaluation. Unless otherwise stated, we also keep the remaining fine-tuning and evaluation settings consistent with \cref{app:exp_details_ftaas}. To reduce compute, we report results using a single random seed.

\subsection{Robust Unlearning against Relearning Attack}
\label{app:exp_details_unlearning}

\textbf{Unlearning setup.} We compare two methods:
\begin{itemize}[leftmargin=15pt]
    \item \textbf{NPO}: Negative Preference Optimization (NPO) is an unlearning method that discourages the model from generating the forget set examples by treating them as negative preferences.
    The total objective is minimized as 
    \begin{align}
    \mathcal{L} = \mathcal{L}_{\text{NPO}} + \gamma \mathcal{L}_{\text{retain}},
    \end{align}
    where $\mathcal{L}_{\text{NPO}} = - \log \sigma \left( - \beta \log \frac{\pi_{\theta}(y|x)}{\pi_{\text{ref}}(y|x)} \right)$ is the NPO loss on the forget set, and $\mathcal{L}_{\text{retain}}$ is the cross-entropy loss on the retain set.
    For hyperparameters, we use $\beta=0.0225$, $\gamma=1.0$, learning rate $5 \times 10^{-6}$, and fine-tune for 250 steps with a batch size of 4 and gradient accumulation steps set to 1.
    \item \textbf{NPO+BECON}: This setup is similar to NPO, but with additional regularization. The hyper-parameter strength for $\mathcal{L}_{\text{plateau}}$ is $\lambda_1=1.0$. For $\mathcal{L}_{\text{cn-min}}$ and $\mathcal{L}_{\text{cn-max}}$, we use $\lambda_2=0.005$ and $\lambda_3=0.005$. We set $\rho = \epsilon=0.01$ and $\tau=1.0$. The batch size is 4, sub-minibatch size is 2, and gradient accumulation steps are set to 1.
\end{itemize}

\textbf{Relearning setup.} We use a learning rate of $1 \times 10^{-5}$, batch size of 1, and gradient accumulation steps set to 4. The relearning attack is standard SFT on the forget set\dots

Finally, the results are average over three random seeds.

\section{More Experiments}

\subsection{Experiments on More Recent Models}
\label{app:exp_qwen3}

We used Llama-2-7B \citep{touvron2023llama}, Qwen-2-7B \citep{yang2024qwen2}, and Gemma-2-9B \citep{team2024gemma} as the model architectures in our main experiments to enable direct comparison with prior work \citep{huang2024vaccine, huang2024lisa, huang2025booster}. To further evaluate whether our findings hold for more recent models, we conduct additional experiments on Qwen-3-4B \citep{yang2025qwen3} under the FTaaS setting following the setup in \cref{sec:ftaas}.

\begin{table*}[!h]
\centering
\caption{Performance of Qwen-3-4B \citep{yang2025qwen3} trained on different fine-tuning datasets in the FTaaS setting. The best and the second best are highlighted in \colorbox[RGB]{220, 235, 255}{blue} and \secondbest{gray}, respectively.}
\label{tab:experiment_recent_model}
\begin{adjustbox}{max width=0.8\textwidth}
\begin{tabular}{l|cc|cc|cc}
\toprule
Methods & \multicolumn{2}{c|}{\textbf{SST2}} & \multicolumn{2}{c|}{\textbf{AGNEWS}} & \multicolumn{2}{c}{\textbf{GSM8K}} \\
\cmidrule(lr){2-3} \cmidrule(lr){4-5} \cmidrule(lr){6-7}
 & HS $\downarrow$ & FA $\uparrow$ & HS $\downarrow$ & FA $\uparrow$ & HS $\downarrow$ & FA $\uparrow$ \\
\midrule
SFT          & 17.45 & 93.23 & 15.35 & \secondbest{83.20} & 7.77 & 65.53 \\
Booster      & \secondbest{3.96} & \secondbest{93.50} & \secondbest{2.10} & \best{83.27} & \secondbest{6.91} & \best{70.80} \\
BECON        & \best{2.67} & \best{93.54} & \best{1.33} & 82.13 & \best{4.77} & \secondbest{69.57} \\
\bottomrule
\end{tabular}
\end{adjustbox}
\end{table*}

As shown in Table~\ref{tab:experiment_recent_model}, BECON continues to achieve the strongest Harmful Score results while maintaining competitive Fine-tuning Accuracy.

\subsection{Full Fine-Tuning Experiments}
\label{app:full_finetuning}

Most experiments in this work use LoRA \citep{hu2022lora}, following prior defenses against harmful fine-tuning \citep{huang2024lisa, huang2024vaccine, huang2025booster}, as LoRA is the most common and practical setting in both FTaaS and open-weight fine-tuning. During immunization, however, the learned LoRA update is merged into the base model before release. The released checkpoint is therefore a standard dense model whose immunization is incorporated into its model parameters and cannot be directly removed by discarding an adapter. This makes LoRA a suitable and efficient mechanism for the immunization stage, while still allowing post-release adversaries to attack the resulting dense model using either LoRA or full-parameter fine-tuning.

\begin{table}[!h]
\centering
\caption{Harmful immunization performance of Llama-2-7B under full-parameter fine-tuning across different learning rates. The number of harmful samples is 100.}
\label{tab:full_finetuning_varying_learning_rate}
\begin{adjustbox}{max width=0.7\columnwidth}
\begin{tabular}{l|ccccc}
\toprule
Methods & \multicolumn{5}{c}{\textbf{Harmful Score} $\downarrow$} \\
\cmidrule(lr){2-6}
 & lr=1e-6 & lr=2e-6 & lr=5e-6 & lr=7e-6 & lr=1e-5 \\
\midrule
BECON & 0.86 & 0.72 & 0.86 & 1.14 & 4.86 \\
\bottomrule
\end{tabular}
\end{adjustbox}
\end{table}

\begin{table}[!h]
\centering
\caption{Harmful immunization performance of Llama-2-7B under full-parameter fine-tuning across different numbers of epochs. The number of harmful samples is 100.}
\label{tab:full_finetuning_varying_epochs}
\begin{adjustbox}{max width=0.7\columnwidth}
\begin{tabular}{l|ccccc}
\toprule
Methods & \multicolumn{5}{c}{\textbf{Harmful Score} $\downarrow$} \\
\cmidrule(lr){2-6}
 & Epoch=1 & Epoch=2 & Epoch=3 & Epoch=4 & Epoch=5 \\
\midrule
BECON & 0.86 & 1.57 & 1.72 & 3.43 & 5.87 \\
\bottomrule
\end{tabular}
\end{adjustbox}
\end{table}

To evaluate full-parameter fine-tuning attacks after model release, we conduct additional experiments on the LoRA-immunized Llama-2-7B model. We use 100 harmful samples and otherwise follow the open-weight immunization setup in \cref{sec:open_weight_immunization}. As shown in \cref{tab:full_finetuning_varying_learning_rate,tab:full_finetuning_varying_epochs}, BECON maintains low Harmful Score across both learning-rate and epoch sweeps, indicating that the immunized dense model remains robust even when the attacker updates all model parameters.

\subsection{Additional Baselines in the Open-Weight Setting}
\label{app:open_weight_more_baselines}

In \cref{sec:open_weight_immunization}, we compare BECON with TAR and RepNoise because both were designed for the open-weight setting. Since Vaccine \citep{huang2024vaccine} and Booster \citep{huang2025booster} are also alignment-stage defenses that act before the model is released, we further evaluate them under the same open-weight attack protocol, sweeping the attacker's number of fine-tuning epochs and learning rate as in \cref{tab:experiment_varying_epochs,tab:experiment_varying_learning_rate}.

\begin{table}[!h]
\centering
\caption{Harmful immunization performance across training epochs, extending \cref{tab:experiment_varying_epochs} with Vaccine and Booster. The top three results in each column are highlighted in bold.}
\label{tab:open_weight_more_baselines_epochs}
\begin{adjustbox}{max width=\columnwidth}
\begin{tabular}{l|cccccc}
\toprule
    Methods & \multicolumn{6}{c}{\textbf{Harmful Score} $\downarrow$} \\
    \cmidrule(lr){2-7}
    & Epoch=5 & Epoch=10 & Epoch=20 & Epoch=50 & Epoch=75 & Epoch=100 \\
\midrule
TAR      & 42.63 & 50.64 & 54.51 & 55.08 & 55.79 & \textbf{54.94} \\
RepNoise & 13.02 & 16.31 & \textbf{15.74} & \textbf{43.35} & \textbf{53.51} & 55.79 \\
Vaccine  & \textbf{1.57} & \textbf{8.30} & 30.04 & 55.08 & 56.94 & 59.37 \\
Booster  & \textbf{1.43} & \textbf{2.58} & \textbf{24.32} & \textbf{48.78} & \textbf{49.21} & \textbf{51.93} \\
BECON    & \textbf{1.00} & \textbf{3.43} & \textbf{27.47} & \textbf{50.21} & \textbf{50.50} & \textbf{52.50} \\
\bottomrule
\end{tabular}
\end{adjustbox}
\end{table}

\begin{table}[!h]
\centering
\caption{Harmful immunization performance across different learning rates, extending \cref{tab:experiment_varying_learning_rate} with Vaccine and Booster. The top three results in each column are highlighted in bold.}
\label{tab:open_weight_more_baselines_lr}
\begin{adjustbox}{max width=\columnwidth}
\begin{tabular}{l|cccccc}
\toprule
    Methods & \multicolumn{6}{c}{\textbf{Harmful Score} $\downarrow$} \\
    \cmidrule(lr){2-7}
    & lr=5e-6 & lr=1e-5 & lr=5e-5 & lr=1e-4 & lr=5e-4 & lr=1e-3 \\
\midrule
TAR      & 28.47 & 42.63 & 56.08 & 55.79 & \textbf{54.08} & \textbf{54.94} \\
RepNoise & 10.16 & 13.02 & \textbf{9.63} & 54.51 & \textbf{52.36} & \textbf{56.08} \\
Vaccine  & \textbf{0.86} & \textbf{1.57} & 35.62 & \textbf{54.36} & 55.94 & 57.80 \\
Booster  & \textbf{1.00} & \textbf{1.43} & \textbf{34.48} & \textbf{49.07} & 55.65 & 56.22 \\
BECON    & \textbf{1.14} & \textbf{1.00} & \textbf{34.33} & \textbf{47.93} & \textbf{54.79} & \textbf{54.51} \\
\bottomrule
\end{tabular}
\end{adjustbox}
\end{table}

As shown in \cref{tab:open_weight_more_baselines_epochs,tab:open_weight_more_baselines_lr}, BECON is the only method that ranks among the top three in every epoch and learning-rate setting. Booster is the strongest additional baseline. It attains a lower harmful score than BECON in five of the six epoch settings, by at most 3.15 points, while BECON attains a lower harmful score than Booster in five of the six learning-rate settings, and Booster falls out of the top three at learning rates of $5\times10^{-4}$ and $10^{-3}$. Vaccine is competitive under small attack budgets but has the highest harmful score among all methods under the largest ones, reaching 59.37 after 100 epochs and 57.80 at a learning rate of $10^{-3}$. Like the other defenses, BECON becomes less effective as the attack budget grows, which motivates future work on more robust defenses in this challenging regime.

\subsection{Seed Variance in the Open-Weight Setting}
\label{app:open_weight_seed_variance}

The open-weight results in \cref{sec:open_weight_immunization} are reported for a single random seed. Since RepNoise attains a lower harmful score than BECON at 20 epochs and at a learning rate of $5\times10^{-5}$ in \cref{tab:experiment_varying_epochs,tab:experiment_varying_learning_rate}, we rerun both methods with three random seeds at these two attack configurations to assess whether this ordering is stable.

\begin{table}[!h]
\centering
\caption{Harmful score of RepNoise and BECON across three random seeds in the open-weight setting of \cref{sec:open_weight_immunization}, at the two attack configurations where RepNoise outperforms BECON in \cref{tab:experiment_varying_epochs,tab:experiment_varying_learning_rate}. Lower is better.}
\label{tab:open_weight_seed_variance}
\begin{adjustbox}{max width=\columnwidth}
\begin{tabular}{l|cccc|cccc}
\toprule
Method & \multicolumn{4}{c|}{\textbf{Epoch=20}} & \multicolumn{4}{c}{\textbf{lr=5e-5}} \\
\cmidrule(lr){2-5} \cmidrule(lr){6-9}
& First seed & Second seed & Third seed & Mean $\pm$ std & First seed & Second seed & Third seed & Mean $\pm$ std \\
\midrule
RepNoise & 15.74 & 44.35 & 36.91 & 32.33 $\pm$ 14.84 & 9.63 & 46.21 & 42.20 & 32.68 $\pm$ 20.06 \\
BECON    & 27.47 & 24.61 & 24.57 & 25.55 $\pm$ 1.66 & 34.33 & 33.05 & 32.90 & 33.43 $\pm$ 0.79 \\
\bottomrule
\end{tabular}
\end{adjustbox}
\end{table}

As shown in \cref{tab:open_weight_seed_variance}, RepNoise is much less stable across seeds than BECON, with standard deviations of 14.84 and 20.06 compared with 1.66 and 0.79 for BECON. The ordering in the main tables comes from the first seed, which is the only seed in which RepNoise attains a lower harmful score than BECON in either configuration. In the other two seeds, BECON attains a lower harmful score in both configurations. Averaged over seeds, BECON has a lower harmful score at 20 epochs (25.55 vs.\ 32.33), while the two methods are comparable at a learning rate of $5\times10^{-5}$ (33.43 vs.\ 32.68).

\subsection{Robustness to Harmful Data Distribution Shift}
\label{app:harmful_data_shift}

Prior experiments in \cref{sec:ftaas} attack with harmful samples drawn from the same BeaverTails distribution as the immunization data, while ensuring the attack set is disjoint from the immunization set. Here, we additionally evaluate robustness under a distribution shift in the harmful data by replacing the attack samples with PureBad-100 \citep{qi2024finetuning}. We keep the rest of the FTaaS setup unchanged and use a harmful ratio of \(p=0.1\).

\begin{table*}[!h]
\centering
\caption{Performance of defense methods on different fine-tuning datasets with a harmful ratio of \(0.1\), where harmful attack samples are drawn from PureBad-100 \citep{qi2024finetuning}. The setup otherwise follows \cref{sec:ftaas}. The best results are highlighted in bold.}
\label{tab:harmful_data_shift_purebad}
\begin{adjustbox}{max width=0.9\textwidth}
\begin{tabular}{l|cc|cc|cc}
\toprule
Method (Llama-2-7B) & \multicolumn{2}{c|}{\textbf{SST2}} & \multicolumn{2}{c|}{\textbf{AGNEWS}} & \multicolumn{2}{c}{\textbf{GSM8K}} \\
\cmidrule(lr){2-3} \cmidrule(lr){4-5} \cmidrule(lr){6-7}
& HS $\downarrow$ & FA $\uparrow$ & HS $\downarrow$ & FA $\uparrow$ & HS $\downarrow$ & FA $\uparrow$ \\
\midrule
SFT & 15.79 & 92.78 & 14.50 & 86.47 & 11.30 & 11.63 \\
BECON (Ours) & \textbf{2.73} & \textbf{93.00} & \textbf{2.96} & \textbf{87.70} & \textbf{2.48} & \textbf{15.33} \\
\bottomrule
\end{tabular}
\end{adjustbox}
\end{table*}

As shown in \autoref{tab:harmful_data_shift_purebad}, BECON remains robust when the harmful attack data comes from a different distribution. Across SST2, AGNEWS, and GSM8K, it substantially reduces Harmful Score relative to standard fine-tuning while preserving, and in these experiments slightly improving, downstream task accuracy.

\section{System Evaluation}
\label{app:system_evaluation}

We provide an additional system-level comparison to clarify the computational overhead of the methods evaluated in \cref{sec:open_weight_immunization}. Here, efficiency refers to making condition-number-based immunization practical in the large-scale LLM setting, including the estimation of Hessian singular-value surrogates, rather than claiming uniformly lower overhead than prior harmful fine-tuning defenses.

\begin{table*}[!h]
\centering
\caption{System evaluation for methods in \cref{sec:open_weight_immunization} on Llama-2-7B. Lower is better for all metrics.}
\label{tab:system_evaluation}
\begin{adjustbox}{max width=0.9\textwidth}
\begin{tabular}{l|ccc}
\toprule
Method (Llama-2-7B) & Clock Time (Hours) $\downarrow$ & Peak Reserved (GB) $\downarrow$ & Peak Allocated (GB) $\downarrow$ \\
\midrule
TAR & 2.7 & 78.07 & 61.06 \\
RepNoise & 2.6 & 87.13 & 65.89 \\
BECON & 5.7 & 59.96 & 51.71 \\
\bottomrule
\end{tabular}
\end{adjustbox}
\end{table*}

As shown in \cref{tab:system_evaluation}, TAR and RepNoise run faster than BECON, but they require substantially more GPU memory. In particular, TAR and RepNoise reach 2.7 and 2.6 hours, respectively, while BECON requires 5.7 hours. However, BECON reduces peak reserved memory to 59.96 GB, compared with 78.07 GB for TAR and 87.13 GB for RepNoise, and likewise reduces peak allocated memory to 51.71 GB, compared with 61.06 GB and 65.89 GB.

This trade-off is consistent with how the methods are implemented. TAR and RepNoise retain hidden states to compute their losses, which increases memory usage. BECON instead keeps only a single copy of the model and reuses gradients to recover the model between perturbations, reducing memory usage at the cost of additional wall-clock time. More broadly, curvature-based immunization remains computationally demanding because the exact Hessian condition number is infeasible to compute for LLMs, and although BECON's approximations in \cref{sect:method} reduce GPU memory relative to the condition-number framework in \citep{zheng2025model} (59.96 GB vs.\ 79.53 GB), they require a longer runtime (\cref{tab:condition_number_system}). Moreover, this overhead is a \emph{one-time cost} incurred during the immunization stage before the model is released, and we leave further efficiency improvements to future work.

\section{Further Discussion with Prior Works}
\label{app:further_discussion}

BECON is related to two lines of prior work, condition-number-based immunization~\citep{zheng2025model} and the attenuation of harmful perturbations in Booster~\citep{huang2025booster}. Our contribution is not the union of their objectives, but the result of \cref{thm:main} that the two components must be combined, together with a realization of condition-number control that is usable at LLM scale. BECON therefore implements this analysis rather than combining existing objectives post hoc. Neither line of prior work predicts the necessity of the other. \citet{zheng2025model} certify slow harmful fine-tuning only through an upper bound, while Booster provides no curvature or convergence analysis and makes no claim about the Hessian spectrum. Moreover, a direct adaptation of \citet{zheng2025model} to LLM immunization does not reduce the harmful score relative to SFT (\cref{tab:condition_number_prior_framework,tab:experiment_different_model_dataset}) and requires more GPU memory than BECON (\cref{tab:condition_number_system}).

\textbf{Discussion with Booster.}
Booster~\citep{huang2025booster} is an alignment-stage defense that attenuates harmful perturbations by minimizing the reduction in harmful loss after a small harmful update. BECON's harmful plateau promotion has a closely related form: it also discourages a local decrease in harmful loss after moving in a harmful update direction. The distinction is primarily in the role this term plays in the full method. In Booster, attenuating harmful loss reduction is the central objective. In BECON, harmful plateau promotion is introduced as a complement to condition-number enlargement, motivated by the theoretical observation that ill conditioning alone does not necessarily prevent rapid harmful optimization when harmful gradients remain large. This motivation comes from the lower bound in \cref{thm:main}, in which small harmful gradient norms are required in addition to a large condition number, rather than from the heuristic of attenuating harmful loss reduction. Reaching a closely related formulation from this principle explains why the term is needed and what it must be combined with. Thus, the plateau term suppresses locally harmful gradients while the condition-number terms distort the harmful curvature geometry.

\begin{table}[!h]
\centering
\caption{Effect of the harmful plateau promotion term relative to Booster in the FTaaS setting on Llama2-7B. Results are averaged over three random seeds. HS denotes Harmful Score, and FA denotes Fine-tuning Accuracy.}
\label{tab:plateau_weighting_booster}
\setlength{\tabcolsep}{5pt}
\begin{tabular}{l|cc|cc|cc}
\toprule
Method & \multicolumn{2}{c|}{SST2} & \multicolumn{2}{c|}{AGNEWS} & \multicolumn{2}{c}{GSM8K} \\
\cmidrule(lr){2-3} \cmidrule(lr){4-5} \cmidrule(lr){6-7}
 & HS $\downarrow$ & FA $\uparrow$ & HS $\downarrow$ & FA $\uparrow$ & HS $\downarrow$ & FA $\uparrow$ \\
\midrule
Booster & 14.31 & 92.59 & \textbf{15.88} & \textbf{86.70} & 9.06 & \textbf{16.27} \\
BECON (\emph{harmful plateau only}) & \textbf{13.07} & \textbf{92.81} & 15.93 & 86.67 & \textbf{8.92} & 15.30 \\
\bottomrule
\end{tabular}
\end{table}

\Cref{tab:plateau_weighting_booster} compares Booster with the BECON variant that keeps only harmful plateau promotion and disables the condition-number terms. This isolates the practical effect of BECON's weighted plateau objective from the full condition-number component. The weighted plateau variant slightly improves harmful score on SST2 and GSM8K while maintaining comparable downstream fine-tuning accuracy. On AGNEWS, the two methods are essentially tied in safety and utility. These results suggest that the plateau weighting can be useful, while the strongest gains in BECON come from combining plateau promotion with condition-number enlargement.

\textbf{Comparison with Booster on more recent models.} We further compare BECON with Booster on Llama-3.1-8B~\citep{grattafiori2024llama} and Qwen3-8B~\citep{yang2025qwen3} under the FTaaS setting of \cref{sec:ftaas}. As shown in \cref{tab:booster_recent_models}, BECON outperforms Booster in both harmful score and fine-tuning accuracy on all three datasets for both models, although the gain is small in some cases.

\begin{table}[!h]
\centering
\caption{Comparison between Booster and BECON on more recent models in the FTaaS setting. Results are reported as mean $\pm$ standard deviation. HS denotes Harmful Score, and FA denotes Fine-tuning Accuracy. The best results for each model are highlighted in bold.}
\label{tab:booster_recent_models}
\begin{adjustbox}{max width=\textwidth}
\begin{tabular}{ll|cc|cc|cc}
\toprule
Model & Method & \multicolumn{2}{c|}{SST2} & \multicolumn{2}{c|}{AGNEWS} & \multicolumn{2}{c}{GSM8K} \\
\cmidrule(lr){3-4} \cmidrule(lr){5-6} \cmidrule(lr){7-8}
 & & HS $\downarrow$ & FA $\uparrow$ & HS $\downarrow$ & FA $\uparrow$ & HS $\downarrow$ & FA $\uparrow$ \\
\midrule
Llama-3.1-8B & Booster & 11.26 $\pm$ 4.59 & 93.23 $\pm$ 0.34 & 6.34 $\pm$ 0.65 & 83.80 $\pm$ 0.62 & 7.87 $\pm$ 3.78 & 45.50 $\pm$ 1.04 \\
 & BECON & \textbf{6.16} $\pm$ 0.40 & \textbf{93.92} $\pm$ 0.49 & \textbf{4.94} $\pm$ 1.32 & \textbf{84.20} $\pm$ 0.14 & \textbf{4.79} $\pm$ 0.91 & \textbf{45.75} $\pm$ 1.34 \\
\midrule
Qwen3-8B & Booster & 3.00 $\pm$ 0.76 & 93.65 $\pm$ 0.46 & 2.19 $\pm$ 0.46 & 83.17 $\pm$ 0.75 & 1.95 $\pm$ 0.36 & 72.63 $\pm$ 0.45 \\
 & BECON & \textbf{2.29} $\pm$ 0.66 & \textbf{93.69} $\pm$ 0.41 & \textbf{1.29} $\pm$ 0.38 & \textbf{84.17} $\pm$ 0.68 & \textbf{1.62} $\pm$ 0.46 & \textbf{73.40} $\pm$ 0.44 \\
\bottomrule
\end{tabular}
\end{adjustbox}
\end{table}

\textbf{Discussion on GSM8K.} In \cref{tab:experiment_different_model_dataset}, Booster attains higher fine-tuning accuracy than BECON on GSM8K for all three models and a lower harmful score on Qwen2-7B. On Llama-3.1-8B and Qwen3-8B, however, BECON outperforms Booster on GSM8K in both metrics (\cref{tab:booster_recent_models}). We therefore conjecture that the gap in \cref{tab:experiment_different_model_dataset} mainly comes from our use of a single default hyperparameter configuration across all fine-tuning datasets, which may not be optimal for GSM8K. Since GSM8K is a mathematical reasoning dataset that differs more from SST2 and AGNEWS than these two differ from each other, dataset-specific factors may also contribute, and we leave the study of such defenses on reasoning models and datasets to future work.

\textbf{Discussion with Antibody.}
Both Antibody~\citep{nguyen2026antibody} and BECON use weighting mechanisms to account for the fact that harmful samples do not contribute equally during optimization. Antibody combines an alignment-stage flatness objective with a fine-tuning-stage weighting mechanism that reduces harmful gradient influence during downstream fine-tuning. BECON is closest to Antibody's alignment-stage component in that both encourage a flatter harmful loss landscape before the model is fine-tuned. Both thus relate to the literature on Sharpness-Aware Minimization~\citep{foret2021sharpnessaware, le-etal-2024-sharpseq, dung2024sharpnessaware, luo2025unveiling}. The main difference is where the weighting is applied and what signal defines the weights. BECON incorporates per-sample weighting directly into the immunization objective. Its weights are computed from the sensitivity of each harmful sample to the plateau perturbation, so the alignment stage emphasizes harmful samples whose losses are most affected by the local harmful update. In contrast, Antibody's fine-tuning-stage weights compare safe and target response likelihoods to reduce harmful gradient influence while preserving learning on benign task data.

\begin{table}[!h]
\centering
\caption{Comparison with Antibody's alignment-stage solution in the FTaaS setting on Llama2-7B and GSM8K.}
\label{tab:antibody_alignment_comparison}
\begin{tabular}{lcc}
\toprule
Method & HS $\downarrow$ & FA $\uparrow$ \\
\midrule
Antibody (\emph{alignment stage only}) & 10.87 & 15.40 \\
BECON & \textbf{3.72} & \textbf{15.43} \\
\bottomrule
\end{tabular}
\end{table}

Because Antibody includes both alignment-stage and fine-tuning-stage components, \Cref{tab:antibody_alignment_comparison} compares BECON against Antibody's alignment-stage solution to isolate the setting in which both methods intervene before downstream fine-tuning. BECON achieves a lower harmful score than Antibody's alignment-stage component on GSM8K while preserving similar fine-tuning accuracy. This comparison supports the view that BECON advances alignment-stage immunization by combining harmful plateau promotion with condition-number regularization, rather than relying on flatness alone.

\section{Safety Loss Landscape}
\label{app:safety_loss_landscape}

\begin{figure}[!h]
\centering
\begin{subfigure}{0.9\linewidth}
\centering
\includegraphics[width=\linewidth]{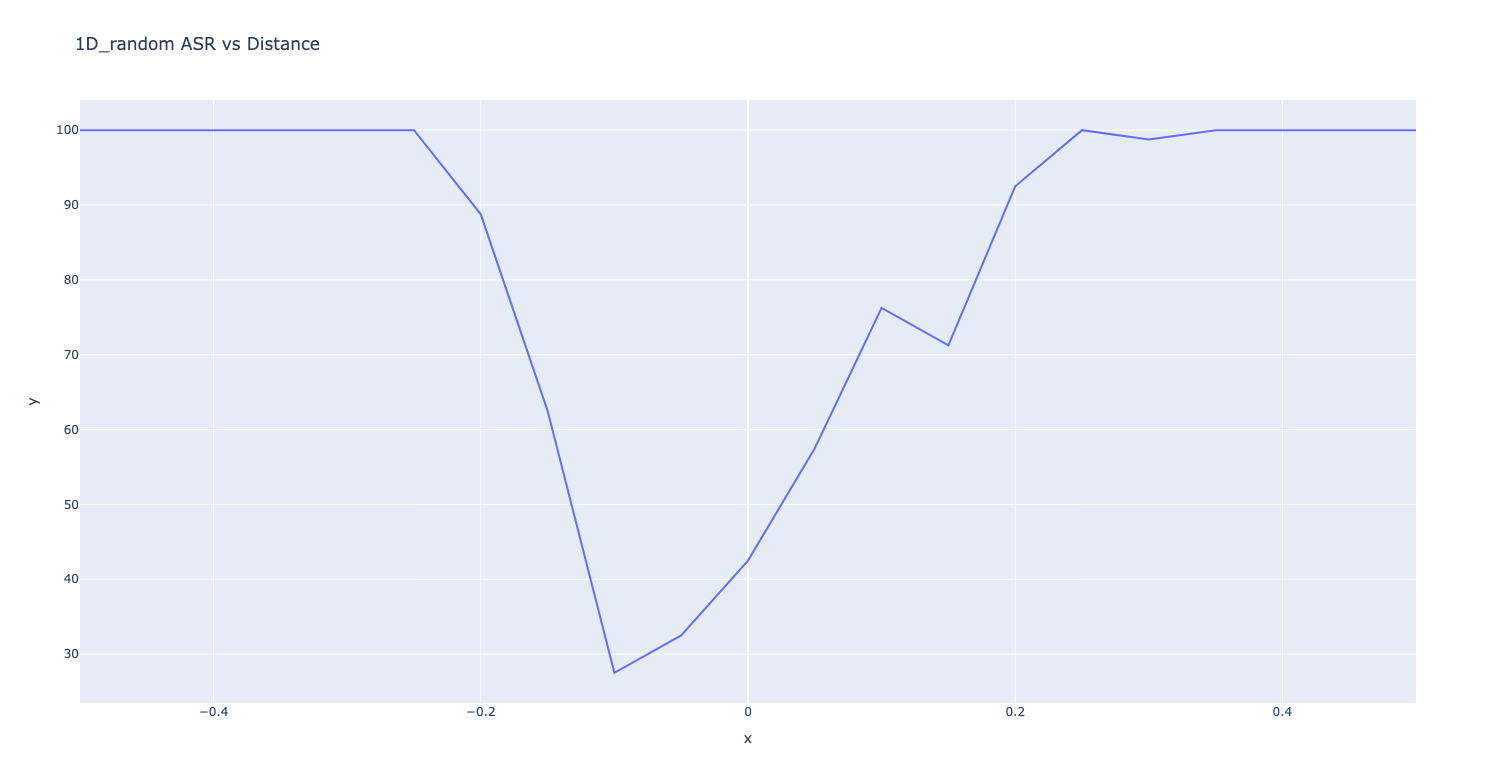}
\caption{BECON}
\end{subfigure}
\vspace{0.5em}
\begin{subfigure}{0.9\linewidth}
\centering
\includegraphics[width=\linewidth]{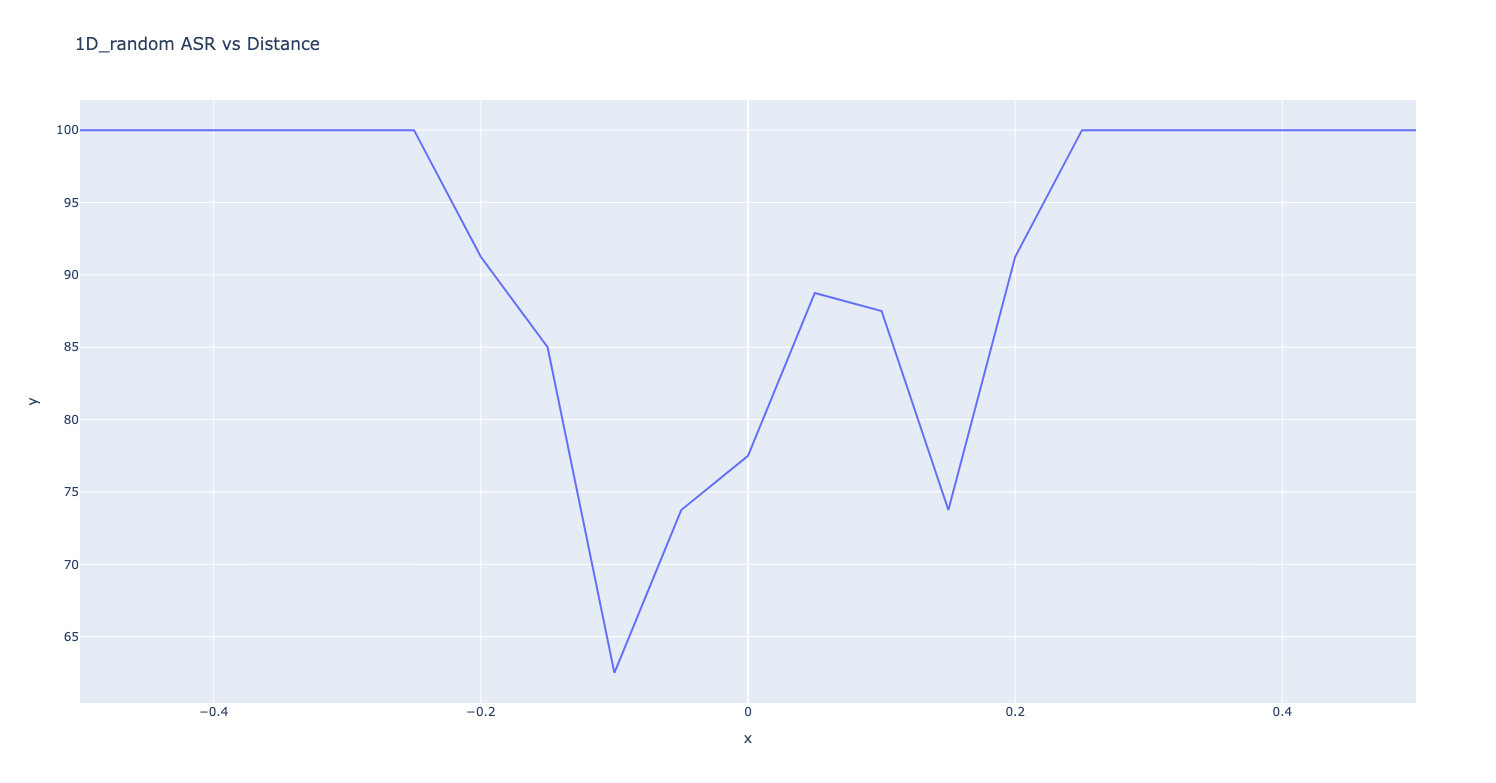}
\caption{Booster}
\end{subfigure}
\caption{1D ASR safety landscapes. Note the smaller y-axis range for BECON compared to Booster.}
\label{fig:visage_landscape}
\end{figure}

\autoref{fig:visage_landscape} visualizes the 1D Attack Success Rate (ASR) landscape of BECON and Booster using the protocol of \citet{peng2024navigating}. The VISAGE score proposed in \citet{peng2024navigating} measures the average safety margin over sampled perturbations in the local safety landscape, with higher values indicating a safer local region around the model. BECON achieves a VISAGE score of 31.81, higher than Booster's 18.75, suggesting that the combination of condition-number regularization and smoothness promotion creates a stronger local safety barrier.

\section{Additional Discussion on Condition-Number Regularization}
\label{app:condition_number_discussion}

This section provides additional discussion of the condition-number component in BECON. A natural question is whether the observed defense effect should be attributed to the intended condition-number-enlarging mechanism, or more simply to suppressing harmful gradients and promoting flatter harmful regions. Our theoretical analysis in the main paper already shows that condition number alone is not sufficient to explain harmful fine-tuning difficulty, which is precisely why BECON combines condition-number regularization with harmful loss plateau promotion. At the same time, condition-number enlargement remains a useful component of the overall method: \textbf{\Cref{tab:condition_number_combined} shows that condition-number regularization alone consistently reduces Harmful Score relative to SFT across models and tasks, although the full BECON objective is usually stronger}.

\begin{table}[!h]
\centering
\caption{Performance of different methods across models. Numbers in parentheses indicate the change in Harmful Score relative to SFT, where negative values indicate improved safety. The harmful score reduction effect of condition number regularization is highlighted in bold.}
\label{tab:condition_number_combined}
\setlength{\tabcolsep}{5pt}
\begin{adjustbox}{max width=0.95\textwidth}
\begin{tabular}{ll|cc|cc|cc}
\toprule
\multicolumn{2}{c|}{\textbf{Model \& Method}} & \multicolumn{2}{c|}{\textbf{SST2}} & \multicolumn{2}{c|}{\textbf{AGNEWS}} & \multicolumn{2}{c}{\textbf{GSM8K}} \\
\cmidrule(lr){3-4} \cmidrule(lr){5-6} \cmidrule(lr){7-8}
\multicolumn{2}{c|}{} & HS $\downarrow$ & FA $\uparrow$ & HS $\downarrow$ & FA $\uparrow$ & HS $\downarrow$ & FA $\uparrow$ \\
\midrule
\multirow{3}{*}{Llama2-7B} & SFT & 36.29 & 92.70 & 34.57 & 85.40 & 23.94 & 10.90 \\
& BECON (\emph{condition-number only}) & 12.49 \textbf{(-23.80)} & 92.70 & 13.59 \textbf{(-20.98)} & 85.93 & 20.13 \textbf{(-3.81)} & 15.00 \\
& BECON & 11.16 (-25.13) & 93.15 & 13.59 (-20.98) & 86.77 & 3.72 (-20.22) & 15.43 \\
\midrule
\multirow{3}{*}{Gemma2-9B} & SFT & 40.77 & 94.11 & 39.91 & 84.47 & 32.05 & 56.73 \\
& BECON (\emph{condition-number only}) & 19.98 \textbf{(-20.79)} & 94.38 & 14.21 \textbf{(-25.70)} & 84.80 & 8.72 \textbf{(-23.33)} & 57.60 \\
& BECON & 16.21 (-24.56) & 94.42 & 4.24 (-35.67) & 84.80 & 8.15 (-23.90) & 58.07 \\
\midrule
\multirow{3}{*}{Qwen2-7B} & SFT & 33.86 & 93.31 & 31.00 & 85.77 & 14.54 & 64.66 \\
& BECON (\emph{condition-number only}) & 7.29 \textbf{(-26.57)} & 93.77 & 7.10 \textbf{(-23.90)} & 87.77 & 3.62 \textbf{(-10.92)} & 67.17 \\
& BECON & 3.34 (-30.52) & 94.76 & 2.91 (-28.09) & 86.00 & 2.48 (-12.06) & 67.93 \\
\midrule
\multirow{3}{*}{Qwen3-4B} & SFT & 17.45 & 93.23 & 15.35 & 83.20 & 7.77 & 65.53 \\
& BECON (\emph{condition-number only}) & 6.01 \textbf{(-11.44)} & 93.81 & 3.96 \textbf{(-11.39)} & 83.87 & 3.05 \textbf{(-4.72)} & 69.70 \\
& BECON & 2.67 (-14.78) & 93.54 & 1.33 (-14.02) & 82.13 & 4.77 (-3.00) & 69.57 \\
\bottomrule
\end{tabular}
\end{adjustbox}
\end{table}

To verify that this reduction is not an artifact of a particular training run, we rerun the condition-number-only variant on Qwen2-7B with three random seeds in the FTaaS setting of \cref{sec:ftaas}. As shown in \cref{tab:condition_number_only_seeds}, the standard deviation across seeds is at most 1.47 points of harmful score, whereas the reduction relative to SFT ranges from 10.92 to 26.57 points. The reduction is therefore at least seven times the seed-to-seed standard deviation on every dataset, which confirms that the condition-number term alone gives a reliable harmful score reduction.

\begin{table}[!h]
\centering
\caption{Harmful score of BECON (\emph{condition-number only}) on Qwen2-7B across three random seeds, reported as mean $\pm$ standard deviation. SFT values are taken from \cref{tab:condition_number_combined}. HS denotes Harmful Score.}
\label{tab:condition_number_only_seeds}
\begin{tabular}{l|ccc}
\toprule
Method (Qwen2-7B) & SST2 HS $\downarrow$ & AGNEWS HS $\downarrow$ & GSM8K HS $\downarrow$ \\
\midrule
SFT & 33.86 & 31.00 & 14.54 \\
BECON (\emph{condition-number only}) & 7.29 $\pm$ 1.03 & 7.10 $\pm$ 0.93 & 3.62 $\pm$ 1.47 \\
\bottomrule
\end{tabular}
\end{table}

Additionally, to clarify why this component is implemented through a surrogate rather than the exact Hessian condition number, it is important to consider the computational constraints of the LLM setting. Directly optimizing the true Hessian condition number is infeasible in large autoregressive models. Computing $\kappa(H)=\sigma_{\max}(H)/\sigma_{\min}(H)$ requires estimating both extremal singular values of a very large Hessian, which is prohibitively expensive in practice. For this reason, some surrogate is unavoidable in large-scale settings. This approximation issue is not unique to our method. Prior condition-number-based immunization frameworks such as \citet{zheng2025model} also rely on nontrivial approximations, since their analysis is developed for linear models with squared loss and their deep-net extension estimates a linear-probing feature-space Hessian, rather than the Hessian over all trainable parameters. In contrast, BECON is developed directly for large autoregressive models and uses the practical surrogate defined in \cref{eq:cn_proxies} to aggregate curvature information across the trainable parameter vector, such as all LoRA parameters, without explicitly computing Hessian singular values.

\begin{figure}[t]
\centering
\includegraphics[width=\linewidth]{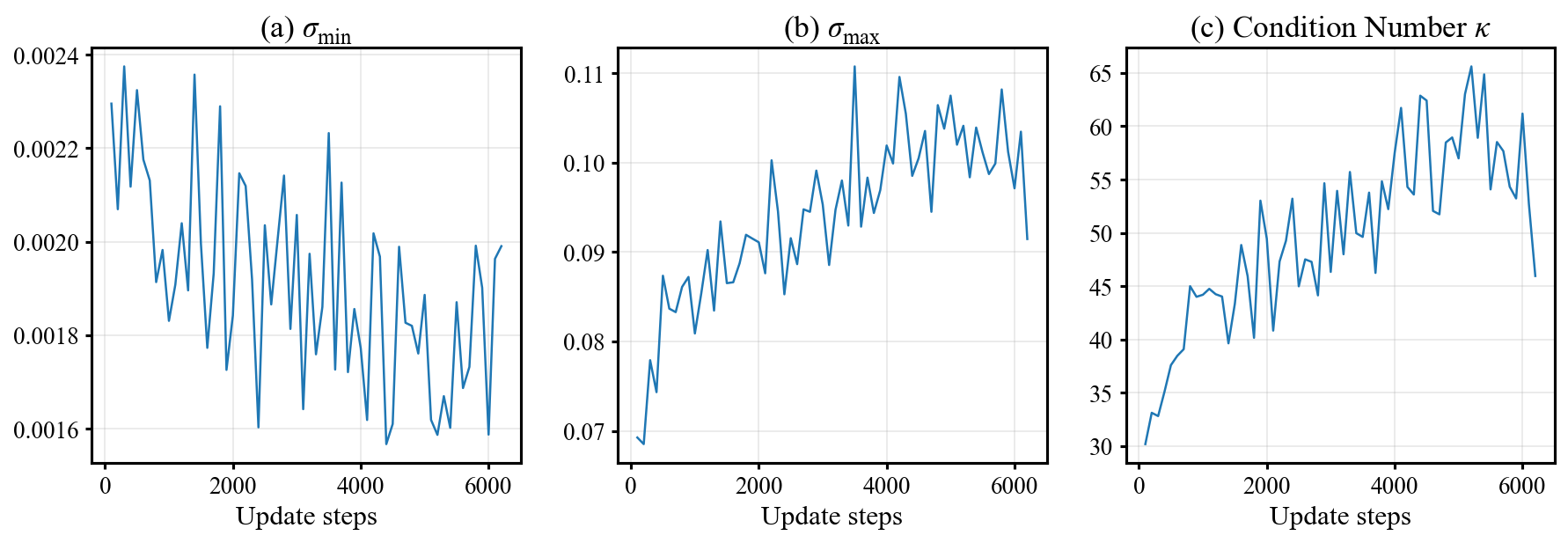}
\caption{Evolution of $\sigma_{\min}$, $\sigma_{\max}$, and the condition number $\kappa = \sigma_{\max}/\sigma_{\min}$ during the immunization process of our method. Due to the substantial computational cost of estimating the condition number, which involves Hessian computations, we conduct this experiment on the Qwen3-0.6B model and visualize the metrics for a single LoRA module in the final block, specifically the LoRA of the $v_{\mathrm{proj}}$ matrix. The decrease in $\sigma_{\min}$ and the increase in $\sigma_{\max}$ together produce the desired effect of an increasing condition number.}
\label{fig:condition_number_trend}
\end{figure}

Although the surrogate defined in \cref{eq:cn_proxies} is not the exact condition number, it is designed to decrease a proxy for $\sigma_{\min}$ and increase a proxy for $\sigma_{\max}$. \Cref{fig:condition_number_trend} visualizes this trend directly. The surrogate does not merely reduce harmful gradients uniformly; it also changes the spectrum in the intended direction by decreasing $\sigma_{\min}$, increasing $\sigma_{\max}$, and thereby enlarging the effective condition number.

\textbf{Small harmful gradients do not imply proximity to a harmful optimum.} The lower bounds in \cref{thm:main} increase as the harmful gradient norms decrease, so BECON promotes small harmful gradients. One may ask whether a small harmful gradient instead indicates that the model is already close to the harmful optimum. This is not the case. BECON creates the plateau at a safety-aligned parameter, not at a harmful optimum. The overall objective always retains the alignment loss $\mathcal{L}_{\text{align}}$ on $\mathcal{D}_{\mathrm{align}}$, which pairs harmful requests with safe responses and anchors the model in the refusal region. The plateau and condition-number terms then flatten and distort the harmful loss landscape around this safe point. A small harmful gradient therefore means that the model sits in a flat region where gradient descent on harmful data has little to follow, not that the harmful task has been learned. Lemma~\ref{PL_inequality}, applied to the harmful loss $\mathcal{L}$, is consistent with this view. Rearranging its two parts gives
\begin{equation}
\frac{\Vert\nabla\mathcal{L}(\bm{\theta})\Vert^{2}}{2\sigma_{\max}} \;\leq\; \mathcal{L}(\bm{\theta})-\mathcal{L}^{*} \;\leq\; \frac{\Vert\nabla\mathcal{L}(\bm{\theta})\Vert^{2}}{2\sigma_{\min}}.
\label{eq:pl_sandwich}
\end{equation}
The right inequality, which follows from Lemma~\ref{PL_inequality}(i), is the only one that bounds the harmful gap from above by the gradient norm, and hence the only one that could certify that a small gradient implies closeness to the harmful optimum. It does so only when $\sigma_{\min}$ is bounded away from zero. The condition-number term of BECON deliberately decreases $\sigma_{\min}$ (\cref{fig:condition_number_trend}), which makes this upper bound loose, so a small harmful gradient is compatible with a large remaining harmful gap. The left inequality, which follows from Lemma~\ref{PL_inequality}(ii), is a lower bound on the harmful gap. A small gradient only makes this lower bound uninformative and never implies that the gap itself is small.

\begin{table}[!h]
\centering
\caption{Harmful score before and after the fine-tuning attack on Llama2-7B in the FTaaS setting of \cref{sec:ftaas}, with GSM8K as the fine-tuning dataset. The pre-attack harmful score is measured right after alignment for SFT and right after immunization for BECON, and is reported as mean $\pm$ standard deviation over three random seeds. Post-attack values are taken from \cref{tab:experiment_different_model_dataset}. HS denotes Harmful Score.}
\label{tab:pre_attack_harmful_score}
\begin{tabular}{l|cc}
\toprule
Method (Llama2-7B) & Pre-attack HS $\downarrow$ & Post-attack HS (GSM8K) $\downarrow$ \\
\midrule
SFT & 0.81 $\pm$ 0.30 & 23.94 \\
BECON & 0.67 $\pm$ 0.22 & 3.72 \\
\bottomrule
\end{tabular}
\end{table}

Empirically, \cref{tab:pre_attack_harmful_score} shows that before any attack, BECON is at least as safe as the SFT model, with a harmful score of 0.67 compared with 0.81. The small harmful gradients of BECON therefore do not come from being close to a harmful solution. After harmful fine-tuning on GSM8K, BECON also remains far safer than SFT (3.72 vs.\ 23.94). Suppressing the harmful gradient norm thus keeps the model at a safety-aligned parameter surrounded by a flat and ill-conditioned harmful landscape, which is the behavior prescribed by \cref{thm:main}.

\textbf{Ground-truth labels versus sampled pseudo-labels.} \Cref{thm:hessian_estimation_ce} builds the Gram matrix from gradients computed with labels sampled from the model's predictive distribution, following Sophia~\citep{liu2024sophia}. In our implementation (\cref{alg:algorithm}), we instead compute these gradients with the ground-truth harmful labels. The implemented object is therefore an empirical-Fisher-type Gram matrix, which is not in general an unbiased estimator of the Gauss--Newton matrix. This substitution is a well-established design choice. Curvature estimates built from per-example gradients on observed labels are common in practice~\citep{NEURIPS2020_d1ff1ec8,NEURIPS2024_f23098fa}, and HELENE~\citep{zhao-etal-2025-helene} explicitly replaces the sampled labels of Sophia's estimator with the true labels. Using ground-truth labels is also deterministic and cheaper than sampling a label at every token, and it evaluates the same harmful-label loss that the attacker optimizes.

\begin{table}[!h]
\centering
\caption{Comparison between the ground-truth-label estimator used in our implementation and the pseudo-label estimator prescribed by \cref{thm:hessian_estimation_ce}, in the FTaaS setting of \cref{sec:ftaas} on Llama2-7B. HS denotes Harmful Score, and FA denotes Fine-tuning Accuracy.}
\label{tab:pseudo_label_estimator}
\setlength{\tabcolsep}{5pt}
\begin{adjustbox}{max width=0.95\textwidth}
\begin{tabular}{l|cc|cc|cc}
\toprule
Method (Llama2-7B) & \multicolumn{2}{c|}{\textbf{SST2}} & \multicolumn{2}{c|}{\textbf{AGNEWS}} & \multicolumn{2}{c}{\textbf{GSM8K}} \\
\cmidrule(lr){2-3} \cmidrule(lr){4-5} \cmidrule(lr){6-7}
& HS $\downarrow$ & FA $\uparrow$ & HS $\downarrow$ & FA $\uparrow$ & HS $\downarrow$ & FA $\uparrow$ \\
\midrule
BECON (\emph{ground-truth labels}, as implemented) & 11.16 & 93.15 & 13.59 & 86.77 & 3.72 & 15.43 \\
BECON (\emph{sampled pseudo-labels}, \cref{thm:hessian_estimation_ce}) & 12.83 & 92.93 & 13.50 & 85.63 & 4.39 & 15.70 \\
\bottomrule
\end{tabular}
\end{adjustbox}
\end{table}

Since the two estimators do not coincide in general, we also evaluate the estimator exactly as prescribed by \cref{thm:hessian_estimation_ce}. This variant samples pseudo-labels token-wise from the model's predictive distribution and keeps every other component of BECON unchanged. As shown in \cref{tab:pseudo_label_estimator}, the two variants are closely comparable in both harmful score and fine-tuning accuracy on all three datasets, with differences of at most 1.7 points in harmful score and 1.1 points in fine-tuning accuracy. These results indicate that the ground-truth-label implementation is a reliable practical approximation of the theory-matched estimator.

\begin{table}[!h]
\centering
\caption{Performance of defense methods on different fine-tuning datasets. The setup is similar to Section 5.1 of the main paper. The best results are highlighted in bold.}
\label{tab:condition_number_prior_framework}
\begin{adjustbox}{max width=0.95\textwidth}
\begin{tabular}{l|cc|cc|cc}
\toprule
Method (Llama2-7B) & SST2 HS & SST2 FA & AGNEWS HS & AGNEWS FA & GSM8K HS & GSM8K FA \\
\midrule
Framework in \citep{zheng2025model} & 38.57 & \textbf{93.58} & 38.01 & 84.33 & 27.75 & 11.53 \\
BECON & \textbf{11.16} & 93.15 & \textbf{13.59} & \textbf{86.77} & \textbf{3.72} & \textbf{15.43} \\
\bottomrule
\end{tabular}
\end{adjustbox}
\end{table}

\begin{table}[!h]
\centering
\caption{System cost of different condition-number-based methods on Llama2-7B.}
\label{tab:condition_number_system}
\begin{tabular}{lcc}
\toprule
Method & Clock Time (hours) & GPU Memory (GB) \\
\midrule
Framework in \citep{zheng2025model} & 1.53 & 79.53 \\
BECON & 5.70 & 59.96 \\
\bottomrule
\end{tabular}
\end{table}

Finally, we also compare BECON against a prior condition-number-based framework adapted to the LLM immunization setting. As shown in \cref{tab:condition_number_prior_framework}, the previous framework is substantially less effective than BECON across all three downstream fine-tuning datasets on Llama2-7B. \Cref{tab:condition_number_system} further shows the corresponding time--memory tradeoff: the adapted prior framework has a shorter clock time, but requires substantially higher peak GPU memory because of its feature-Gram and singular-value computation. \textbf{Although our method has a higher one-time runtime, it achieves better memory efficiency, defense performance, and fine-tuning accuracy.} These results support the claim that BECON's practical surrogate is not only computationally viable, but also more effective in the LLM setting than directly adapting earlier condition-number-based approaches.

\section{Impact Statement}
\label{app:impact_statement}

This paper presents work whose goal is to advance the field of Machine Learning by improving the safety and robustness of Large Language Models against harmful fine-tuning. The proposed immunization framework is intended to reduce the risk that deployed or open-weight models can be adapted to produce harmful behavior. However, documenting the weaknesses of current fine-tuning safeguards may also lower the barrier for adversaries to study and exploit these vulnerabilities. We therefore emphasize that the method is presented to strengthen defenses and encourage responsible deployment and evaluation practices.

\newpage
\section*{NeurIPS Paper Checklist}

\begin{enumerate}

\item {\bf Claims}
    \item[] Question: Do the main claims made in the abstract and introduction accurately reflect the paper's contributions and scope?
    \item[] Answer: \answerYes{}
    \item[] Justification: The claims are accurate.

\item {\bf Limitations}
    \item[] Question: Does the paper discuss the limitations of the work performed by the authors?
    \item[] Answer: \answerYes{}
    \item[] Justification: We include the discussion of limitation in Section \ref{app:Limitations_and_future_work} and computational efficiency in Appendix \ref{app:system_evaluation}.

\item {\bf Theory assumptions and proofs}
    \item[] Question: For each theoretical result, does the paper provide the full set of assumptions and a complete (and correct) proof?
    \item[] Answer: \answerYes{}
    \item[] Justification: We clearly state the assumptions and provide all the proofs in the appendix.

    \item {\bf Experimental result reproducibility}
    \item[] Question: Does the paper fully disclose all the information needed to reproduce the main experimental results of the paper to the extent that it affects the main claims and/or conclusions of the paper (regardless of whether the code and data are provided or not)?
    \item[] Answer: \answerYes{}
    \item[] Justification: We provide full details sufficient to reproduce the main experiment in the appendix.

\item {\bf Open access to data and code}
    \item[] Question: Does the paper provide open access to the data and code, with sufficient instructions to faithfully reproduce the main experimental results, as described in supplemental material?
    \item[] Answer: \answerYes{}
    \item[] Justification: Our code is available at \colorhref{https://github.com/minhquoc0712/BECON}{https://github.com/minhquoc0712/BECON}. The datasets used in this paper are publicly available online.

\item {\bf Experimental setting/details}
    \item[] Question: Does the paper specify all the training and test details (e.g., data splits, hyperparameters, how they were chosen, type of optimizer) necessary to understand the results?
    \item[] Answer: \answerYes{}
    \item[] Justification: We specify the main experimental settings in the main paper and provide full details in Appendix \ref{app:experiment_details}.

\item {\bf Experiment statistical significance}
    \item[] Question: Does the paper report error bars suitably and correctly defined or other appropriate information about the statistical significance of the experiments?
    \item[] Answer: \answerYes{}
    \item[] Justification: We vary hyperparameters and evaluate across diverse attack settings, including different model architectures, datasets, and harmful ratios. The breadth of these experiments provides sufficient evidence for statistical significance. Furthermore, most of the main experiment results are averaged over three random seeds, providing additional confidence in the robustness of our findings.

\item {\bf Experiments compute resources}
    \item[] Question: For each experiment, does the paper provide sufficient information on the computer resources (type of compute workers, memory, time of execution) needed to reproduce the experiments?
    \item[] Answer: \answerYes{}
    \item[] Justification: We report the GPU types in Appendix \ref{app:experiment_details}, and the GPU memory and wall-clock time of BECON and the baselines in Appendix \ref{app:system_evaluation}.
    
\item {\bf Code of ethics}
    \item[] Question: Does the research conducted in the paper conform, in every respect, with the NeurIPS Code of Ethics \url{https://neurips.cc/public/EthicsGuidelines}?
    \item[] Answer: \answerYes{}
    \item[] Justification: Our research fully complies with the NeurIPS Code of Ethics.

\item {\bf Broader impacts}
    \item[] Question: Does the paper discuss both potential positive societal impacts and negative societal impacts of the work performed?
    \item[] Answer: \answerYes{}
    \item[] Justification: Yes, we include the impact statement in Appendix \ref{app:impact_statement}.
    
\item {\bf Safeguards}
    \item[] Question: Does the paper describe safeguards that have been put in place for responsible release of data or models that have a high risk for misuse (e.g., pre-trained language models, image generators, or scraped datasets)?
    \item[] Answer: \answerNA{}
    \item[] Justification: The paper poses no such risks.

\item {\bf Licenses for existing assets}
    \item[] Question: Are the creators or original owners of assets (e.g., code, data, models), used in the paper, properly credited and are the license and terms of use explicitly mentioned and properly respected?
    \item[] Answer: \answerYes{}
    \item[] Justification: We use publicly available datasets and models with proper citation and adhere to their respective licenses.

\item {\bf New assets}
    \item[] Question: Are new assets introduced in the paper well documented and is the documentation provided alongside the assets?
    \item[] Answer: \answerYes{}
    \item[] Justification: We release our code at \colorhref{https://github.com/minhquoc0712/BECON}{https://github.com/minhquoc0712/BECON}.

\item {\bf Crowdsourcing and research with human subjects}
    \item[] Question: For crowdsourcing experiments and research with human subjects, does the paper include the full text of instructions given to participants and screenshots, if applicable, as well as details about compensation (if any)? 
    \item[] Answer: \answerNA{}
    \item[] Justification: The paper does not involve crowdsourcing nor research with human subjects.

\item {\bf Institutional review board (IRB) approvals or equivalent for research with human subjects}
    \item[] Question: Does the paper describe potential risks incurred by study participants, whether such risks were disclosed to the subjects, and whether Institutional Review Board (IRB) approvals (or an equivalent approval/review based on the requirements of your country or institution) were obtained?
    \item[] Answer: \answerNA{}
    \item[] Justification: The paper does not involve crowdsourcing nor research with human subjects.

\item {\bf Declaration of LLM usage}
    \item[] Question: Does the paper describe the usage of LLMs if it is an important, original, or non-standard component of the core methods in this research? Note that if the LLM is used only for writing, editing, or formatting purposes and does \emph{not} impact the core methodology, scientific rigor, or originality of the research, declaration is not required.
    \item[] Answer: \answerNA{}
    \item[] Justification: The core method development in this research does not involve LLMs as any important, original, or non-standard components.

\end{enumerate}

\end{document}